# Deep Recurrent Neural Networks for Time-Series Prediction

Sharat C Prasad, *Member, IEEE*, and Piyush Prasad

*Abstract*— Ability of deep networks to extract high level features and of recurrent networks to perform time-series inference have been studied. In view of universality of one hidden layer network at approximating functions under weak constraints, the benefit of multiple layers is to enlarge the space of dynamical systems approximated or, given the space, reduce the number of units required for a certain error. Traditionally shallow networks with manually engineered features are used, back-propagation extent is limited to one and attempt to choose a large number of hidden units to satisfy the Markov condition is made. In case of Markov models, it has been shown that many systems need to be modeled as higher order. In the present work, we present deep recurrent networks with longer back-propagation through time extent as a solution to modeling systems that are high order and to predicting ahead. We study epileptic seizure suppression electro-stimulator. Extraction of manually engineered complex features and prediction employing them has not allowed small low-power implementations as, to avoid possibility of surgery, extraction of any features that may be required has to be included. In this solution, a recurrent neural network performs both feature extraction and prediction. We prove analytically that adding hidden layers or increasing backpropagation extent increases the rate of decrease of approximation error. A Dynamic Programming (DP) training procedure employing matrix operations is derived. DP and use of matrix operations makes the procedure efficient particularly when using data-parallel computing. The simulation studies show the geometry of the parameter space, that the network learns the temporal structure, that parameters converge while model output displays same dynamic behavior as the system and greater than .99 Average Detection Rate on all real seizure data tried.

*Index Terms*— Recurrent Neural Networks, Multi-layer Neural Networks, Artificial Neural Network, Neural Prosthesis

## I. Introduction

TIME-SERIES are found in many areas including communication, health and finance. Large-scale computer and communication networks generate time-varying metric that characterize the state of the networks. Audio signal arriving at a microphone or radio-frequency signal arriving at a receive antenna are series of values. In the area of health Electrocardiograph, electroencephalogram, etc all are or generate time-series data. End-of-day face-values of different financial instruments are time-series. In all these cases, it is of interest to discern patterns, detect behavior that differs from ordinary and predict future behavior.

Many applications have understandably stringent requirements. We pick implantable epileptic seizure suppression electro-stimulator driven by automatic seizure prediction as such an application and to provide the context for a study of time-series prediction. Epileptic seizure is an abnormality characterized by brief and episodic neuronal synchronous discharges with dramatically increased amplitude.

Electroencephalography (EEG) provides measure of cortical activity with millisecond temporal resolution. Signal representations and processing of EEG recordings has been considered since early days of EEG. Recordings are band-pass filtered (e.g. with a pass band of 1-70*Hz*) and digitized (e.g. at 200 samples/second and 12-bit resolution).

The current gold standard for determining onset times is the trained eye of an encephalographer. Expert encephalographers look for rhythmic activity. Brain activity during a seizure displays a great deal of variation from patient to patient and even from one incident to the next for the same patient.

Implantable neural prosthetic that deliver electrical stimulation on-demand have emerged as a viable and preferred treatment for a portion of the population of seizure patients. Up until recently for reasons of feasibility of implementation, systems normally take the form of bed-side equipment communicating with implanted electrodes.

Size is a consideration for any implantable device. Furthermore both on account of power availability and of damage to surrounding tissue by the heat produced by the implanted device, power consumption is a consideration.

Detection algorithms employ characteristics of or features extracted from EEG signal for detection.

Both feature computation and prediction modules contribute to the overall size and power requirement. Feature computation and prediction employ algorithms required to achieve their objective. Each feature computation may compute a distinct feature and only this feature.

Savings result if same computation element is reused for sub-phases of the same computation phase (e.g. different feature computations). Greater savings result if same computation element is reused across phases (e.g. feature computation and prediction).

Given the variability in signature in EEG of seizure even for the same patient, a rich variety of features has been designed to assist with the task of seizure detection [17][9]. Given the

Date first submitted 8/15/2014.
Sharat C Prasad is Best3 Corp, Saratoga, CA 95070 (e-mail: scprasad@ieee.org).
Piyush Prasad is with the Dept of Biomedical Engineering, University of Washington, ST Louis, MO 65030 (e-mail:piyushsprasad@gmail.com).

consideration of surgery, it is desirable that the implant include support for any feature computation that is or may become necessary[1]. Support for comprehensive set of features is made complicated by the possibility that the optimal set of features may differ for the same patient from one stage of treatment to a following stage.

Given the vector of inputs $\mathbf{u}_k$, vector of features $\mathbf{v}_k$ is computed. System is modeled as having internal state $\mathbf{x}_k$ and transition function $f$ such that $\mathbf{x}_{k+1} = f(\mathbf{x}_k, \mathbf{v}_k)$ and measurement function $h$ such that $\hat{z}_k = h(\mathbf{x}_k, \mathbf{v}_k)$, are defined. In defining the internal state $\mathbf{x}_k$, attempt is made to satisfy the Markov condition $P(z_k | \mathbf{v}_k, \mathbf{v}_{k-1}, \cdots, \mathbf{v}_0) = P(z_k | \mathbf{x}_k, \mathbf{v}_k)$.

As is usual to the type of problem at hand, the parameters of the prediction algorithm may need to be set by a training phase. Also, as elsewhere, there are two options to training - offline and online. A consideration with offline trained algorithms is that as parameters change, the offline training needs to be repeated (or also continuously run in parallel) and the detection algorithm needs to be configured afresh with the new parameters.

Systems employing online training can be considered to continuously incrementally train themselves and therefore able to adapt.

We will refer to frameworks where given the sequence of inputs $U^{[k+1]} = \{\mathbf{u}_1, \mathbf{u}_2, \ldots, \mathbf{u}_{k+1}\}$ and observations $Z^{[k]} = \{\mathbf{z}_1, \mathbf{z}_2, \ldots, \mathbf{z}_k\}$ where each $\mathbf{z}_i$ is $\mathbf{z}_i = h(\mathbf{x}_i, \mathbf{u}_i)$, $h$ is a known – the measurement - function but $\mathbf{x}_i$s are not observable, notation $\hat{\mathbf{x}}_{k+1|k}$ stands for estimate of $\mathbf{x}_{k+1}$ given $U^{[k+1]}$ and $Z^{[k]}$, we first predict $\hat{\mathbf{x}}_{k+1|k}$ and then, when $\mathbf{z}_{k+1}$ is available, compute $\{\hat{\mathbf{x}}_{k+1|k+1}, \hat{\mathbf{x}}_{k|k+1}, \ldots, \hat{\mathbf{x}}_{k-q+2|k+1}\}$ as predict-correct frameworks. The ability of predict-correct frameworks to "go back" and compute an improved estimate $\hat{\mathbf{x}}_{k-q+2|k+1}$ than $\hat{\mathbf{x}}_{k-q+2|k-q+1}$ computed earlier offers the possibility that occurrence of an easier to detect event at instant $k+1$ can be used to learn correct mapping when $\mathbf{z}_{k-q+2}$ were either not available or were not informative.

The problem is one of computing parameters $\mathbf{w} = \{w_i\}_{i=1}^{i=M}$ of a parameterized function $f_\mathbf{w}$ such that when it is applied to the training input data $\mathbf{X} = \{\mathbf{x}_j\}_{j=1}^{j=N}$ some metric computed from values $\hat{\mathbf{Z}} = \{\hat{z}_j = f_\mathbf{w}(\mathbf{x}_j)\}_{j=1}^{j=N}$ returned by the function and supplied output values $\mathbf{Z} = \{z_j\}_{j=1}^{j=N}$ is optimized.

For example, the problem can be expressed as

$$\hat{\mathbf{w}} = \underset{\mathbf{w}}{\operatorname{argmin}} \sum \left( (f_\mathbf{w}(\mathbf{u}_j) - z_j)^2 + \lambda g(\mathbf{w}) \right)$$

$$= \underset{\mathbf{w}}{\operatorname{argmin}} \left( (\mathbf{Uw} - \mathbf{z})^T (\mathbf{Uw} - \mathbf{z}) + \lambda \mathbf{w}^T \mathbf{w} \right). \quad (1)$$

---

[1] Another application area with this characteristic is Deep Space Exploration. Once a explorer is launched with certain, preferably learning, computation engines, these computation engines have to learn to solve the problem even if the characteristics of the problems evolve as explorer travels deeper and deeper into outer space.

Here $\lambda$ is a trade-off constant that trades-off training error minimization for regularization.

As expressed above the problem is called ridge-regression and has a simple closed form solution

$$\hat{\mathbf{w}} = (\mathbf{U}^T \mathbf{U} + \lambda \mathbf{I})^{-1} \mathbf{U}^T \mathbf{y}. \quad (2)$$

The above solution represents an off-line training method and, as introduced above has following assumptions:

- $z_j$ is linearly dependent on $\mathbf{u}_j$,
- $\mathbf{u}_j$ s are i.i.d.,
- components of $\mathbf{u}_j$ are not linearly correlated,
- given training pairs ($\mathbf{u}_j$, $z_j$) are noise free and objective and regularization are as in Ridge Regression.

We know each of these to not hold for Epileptic Seizure prediction. Adaptability and therefore online training is desirable in seizure prediction methods. Only a highly complex feature can hope to achieve linear relation to prediction output. Individual feature vectors in a sequence of feature vectors are not independent. Training data is usually significantly noisy. An objective such as maximizing average detection rate is not maximizing linear combination of distances.

When components of $\mathbf{u}_j$ are not completely orthogonal, methods exist to orthogonalize them [13].

When $\mathbf{u}_j$ s are not i.i.d., 'Time Series' methods consider that $\mathbf{u}_j$ is value at 'time instant' $j$, $\mathbf{u}_{j+1}$ is value at 'time instant' $j+1$ and so on. These methods take covariance explicitly into account.

Assumption of linear relation of $z_j$ to $\mathbf{u}_j$ can be eliminated by employing a Neural Network.

Feedback interconnections within Recurrent Neural Networks impart them the ability to retain information and enable them to perform inference tasks such as Epileptic Seizure Prediction involving time series such as Electroencephalographic data of brain activity.

Deep Neural Networks can automatically extract relevant high-level characteristics making it unnecessary to have to manually engineer characteristic extraction and avoiding loss of performance from characteristics being not matched to task at hand. However, most importantly, automatic feature extraction, addresses the need for adaptability whereby features suited to a patient are used.

However both, the network being deep and network incorporating feedback, add to the difficulty of training the network. Approaches considered for training deep networks include combining and following unsupervised pre-training with supervised training [4]. At core of both of these training algorithms is usually the backpropagation gradient descent.

Among the earliest methods considered for training Recurrent Networks is the Back-Propagation Through Time (BPTT) method [24] that can be considered the analogue of the Back-Propagation method for Feed-Forward Networks. However the BPTT method had been believed to be prone to getting caught in many shallow local minima on the error hyper-surface in the parameter hyper-space. Traditionally, for local minima avoidance, unit activation and/or weight update are made stochastic and the activation or the weight are assigned, rather than the exact computed value, a value drawn from the applicable distribution with the computed values as moments of the distribution.

Furthermore limitations stem from approximating the non-linear dynamics of systems as (locally) linear for purposes of deriving the moments of the posterior state distribution and from approximating the state distributions as Gaussian.

It is to be noted that greater the non-linearity greater are the errors introduced by the linear approximation. In case when process and measurement functions are deep neural networks, if the transformation from external inputs thru the collection of layers to final outputs is considered, the transformation represents a large degree of non-linearity.

Computing the derivatives is tedious and is considered another shortcoming of methods requiring derivatives.

It is needed to address both the eventual and the speed of convergence. The model of a system, that when supplied with some input with stationary statistics, is producing the same time varying output as the system, cannot be considered converged if one-step updates are non-zero.

The main contribution of the present work are (i) Deep Recurrent Neural Network as a unified solution for automatic extraction of relevant features, automatic construction of internal state and prediction for time-series data, (ii) theoretical derivation of the broader space of dynamical systems enabled by both more than 1 hidden layer and back-propagation through time by greater than 1 instant (iii) the Back-Propagation Thru Time-and-Space training method as a solution to hard problem of training deep recurrent networks, (iv) the dynamic programming formulation of the weight computation and (v) small, low-power and adaptive epileptic seizure prediction solution.

In outline, this report presents in Section 2, Deep Recurrent Neural Networks and their Back-propagation Thru Time and Space Training. Section 3 discusses universality of Recurrent Neural Networks and the bound on error in approximation by RNN. Section 5 presents simulation study of training of and prediction using a Deep Recurrent Neural Network for automatically extracted features.

## II. Time-Series and Prediction

### A. Epileptic Seizure

Epileptic seizure is an abnormality characterized by brief and episodic neuronal synchronous discharges with dramatically increased amplitude. Seizures are generated by abnormal synchronization of neurons which is unforeseeable for the patients.

Electroencephalography (EEG) provides measure of cortical activity with millisecond temporal resolution. It is a record of electrical potential generated by the cerebral cortex nerve cells (Figure 1). When epileptic brain activity occurs locally in a region of the brain and it is seen only in a few channels of EEG and, when it happens globally in the brain, it is seen in all channels of EEG.

An example setup for EEG recording [21] may include Ag/AgCl disc electrodes placed in accordance with one of the electrode placement systems (e.g. the 10-20 international electrode placement system [16]). Electrodes may be placed on the scalp, or may be leads that are implanted in the cranium. These may be implanted in or on the cortex at the seizure onset zones. An example is the sensor-neuro-stimulator device from NeuroPace [23] that includes an implantable battery-powered neuro-stimulator and quadripolar depth or strip leads that are implanted in the cranium and are used for both sensing and stimulation.

Signal representations and processing of EEG recordings has been considered since early days of EEG. Recordings are band-pass filtered (e.g. with a pass band of 1-70$Hz$) and digitized (e.g. at 200 samples/second and 12-bit resolution). Representations based on Fourier Transform have been popular. Four characteristic bands - delta (<4 Hz), theta (4-8 Hz), alpha (8-13 Hz) and beta (13-10 Hz) are identified in the Fourier Transform.

The current gold standard for determining onset times is the trained eye of an encephalographer. Expert encephalographers look for rhythmic activity. Epileptiform discharge may be seen in the form of spikes repeating at a frequency of 3-Hz and wave complex in absence seizure (Figure 2). Phase synchronization and changes in it, are believed to accompany seizures. Reported measures representative of seizure also include spectral power, synchronization or amplitude deviates from normal bounds.

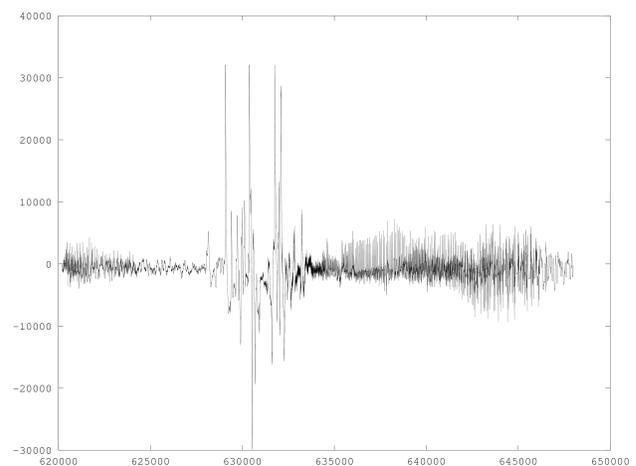

Figure 1: Normal brain activity.

Brain activity during a seizure displays a great deal of variation from patient to patient and even from one incident to the next for the same patient. A patient may exhibit low voltage fast gamma activity, may have a beta frequency burst just prior to high frequency activity, may have onset activity in alpha bands characterized by rhythmic round sinusoidal waves, may have sharp waves in the delta-theta range followed by high amplitude spiking activity, may have one of these onset activities or voltage gamma activity preceded by spike-and-wave activity or may have activity from the above entire range preceded by semi rhythmic slow waves [17].

There are two benefits to the Time Series analysis of EEG data advocated here - (1) offline to determine whether pre-ictal states exist enabling long-term advance prediction of seizure and (2) online employing information gained from (1) for preventative therapy in the form of electro-stimulation that have been seen to be able to suppress or limit the seizure if applied in a timely manner. Electro-stimulation is believed to reset brain dynamic from pre-ictal to inter-ictal state.

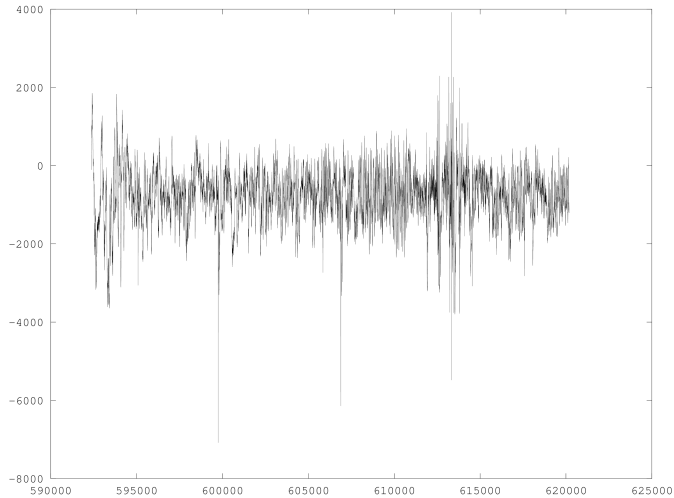

Figure 2: Brain activity during seizure.

Implantable neural prosthetic that deliver electrical stimulation on-demand have emerged as a viable and preferred treatment for a portion of the population of seizure patients. While timely application of electro-stimulation is seen to suppress seizures, extended or more than necessary electro-stimulation is seen to damage neuronal tissue and have such adverse consequences as requiring increasing amounts of stimulation to have effect and stimulation induced suppression of neuronal excitability. Up until recently for reasons of feasibility of implementation, systems normally take the form of bed-side equipment communicating with implanted electrodes. In comparison self-contained neural prosthetic will not require the patient to be bed-bound. Responsive neuro-stimulation, by virtue of being timely and limiting the amount, may also significantly increase the effectiveness of therapy. The advent of very low-power battery-operated electronics has made possible implantable electronics with closed-loop working of detection and stimulation.

Traditionally detection algorithms have emphasized detection efficacy and have disregarded power consumption and size. Size is a consideration for any implantable device. Furthermore both on account of power availability and of damage to surrounding tissue by the heat produced by the implanted device, power consumption is a consideration.

While wireless communication between implanted electrode and external control unit is possible to simplify the implanted electronics, this may limit fine-grained analysis as the possible data rates may remain of the order of 100s of Kbps and therefor too low for some time to come [25]. While new wireless powering technology is coming into being, the rate of energy transferred may remain low for some time to come [18]. Hence size and power consumption need to be considered along side detection efficiency.

Detection algorithms employ characteristics of or features extracted from EEG signal for detection. Feature may be computed on the raw time-domain or frequency or another transform domain representation. For example, sequence of samples can be divided into time windows, frequency transform can be computed for each time window to obtain spectral power in a few different frequency bands, the vector of spectral powers can be a feature vector and a sequence of such vectors may be analyzed by a detection algorithm. Figure 3 shows the overall organization of the detection sub-system comprising of feature computation and prediction modules.

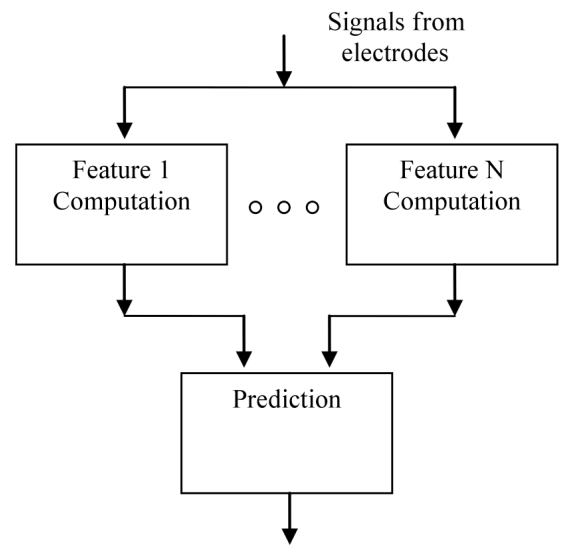

Figure 3: Organization of detection sub-system.

Both feature computation and prediction modules contribute to the overall size and power requirement. Feature computation and prediction employ algorithms required to achieve their objective. Each feature computation may compute a distinct feature and only this feature.

In above, we gave examples of characteristics expert encephalographers look for in the EEG. Some of these characteristics were described in terms of frequency domain representation. To make use of such characteristic, frequency domain representation needs to be computed. Efficient *frequency transform* computation requires specialized algorithms and space, time and power. If realized as circuits, the circuits may not serve to compute a different representation.

Time-domain features avoid the cost of transform computation but require specialized algorithms.

One noteworthy aspect of the transform computation algorithms, that we will use to argue in favor of our thesis, is that their regular structure enables easy to visualize trading-off of space (number of computation engines, amount of memory) for time and remain acceptable as long overall latency remains within limits. To save space and power, computation may be performed iteratively, reusing the same computation element. Savings result if same computation element is reused for sub-phases of the same computation phase (e.g. different feature computations). Greater savings result if same computation element is reused across phases (e.g. feature computation and prediction).

Given the variability in signature in EEG of seizure even for the same patient, a rich variety of features has been designed to assist with the task of seizure detection [17][9]. There have been studies of efficacy of different detection features. Given the consideration of surgery, it is desirable that the implant include support for any feature computation that is or may become necessary[2]. Support for comprehensive set of features is made complicated by the possibility that the optimal set of features may differ for the same patient from one stage of treatment to a following stage.

Advent of small low-power computer-on-chip [20] makes them one option to address the feature computation requirement. However, as numerous studies have shown, size and power requirement of general purpose programmable solutions always exceed custom optimized solutions.

There have also been studies of attempts [9] to improve performance by using detection features in combination. Latter has shown that when all of a set of detection features are used best performance results. However the benefit from using more than a certain number of features may not justify their incremental cost when size and power constraints exist [17].

*B. Time-series prediction*

Given the vector of inputs $\mathbf{u}_k$, vector of features $\mathbf{v}_k$ is computed. System is modeled as having internal state $\mathbf{x}_k$ and transition function $f$ such that $\mathbf{x}_{k+1} = f(\mathbf{x}_k, \mathbf{v}_k)$ and measurement function $h$ such that $\hat{z}_k = h(\mathbf{x}_k, \mathbf{v}_k)$, are defined. In defining the internal state $\mathbf{x}_k$, attempt is made to satisfy the Markov condition $P(z_k | \mathbf{v}_k, \mathbf{v}_{k-1}, \cdots, \mathbf{v}_0) = P(z_k | \mathbf{x}_k, \mathbf{v}_k)$.

As is usual to the type of problem at hand, the parameters of the prediction algorithm may need to be set by a training phase. Also, as elsewhere, there are two options to training - offline and online. An offline (batch) training algorithm is given feature vector $x_i$ and target $z_i$ pairs $\{(x_i, z_i)\}_{i=1}^{i=N}$. The training goes over all the data to determine detection parameters. A consideration with offline trained algorithms is that as parameters change, the offline training needs to be repeated (or also continuously run in parallel) and the detection algorithm needs to be configured afresh with the new parameters.

On the other hand, an online (incremental) algorithm initializes its parameters to certain initial values, is given either just one feature vector $\mathbf{x}_i$ or one feature-vector-target pair $(\mathbf{x}_i, z_i)$. An online algorithm only produces a prediction $\hat{z}_i$ if only given a feature vector or also updates its parameters if given a feature-vector-target pair $(\mathbf{x}_i, z_i)$. To start with, an on-line trained algorithm may produce poor predictions. As the data characteristics are learnt, predictions improve. Systems employing online training can be considered to continuously incrementally train themselves and therefore able to adapt. One issue with online training is that for supervised training, target $z_i$ corresponding to each feature-vector $\mathbf{x}_i$ is required. A second issue is that, on one hand, recent feature-vector-target pairs may cause invariant characteristics to be over-written by transient characteristics and on the other hand, recent pairs may not cause transient characteristics to be overwritten by new transient characteristics.

We will refer to frameworks where given the sequence of inputs $U^{[k+1]} = \{\mathbf{u}_1, \mathbf{u}_2, \ldots, \mathbf{u}_{k+1}\}$ and observations $Z^{[k]} = \{\mathbf{z}_1, \mathbf{z}_2, \ldots, \mathbf{z}_k\}$ where each $\mathbf{z}_i$ is $\mathbf{z}_i = h(\mathbf{x}_{i-1}, \mathbf{u}_i)$, $h$ is a known – the measurement - function but $\mathbf{x}_i$s are not observable, notation $\hat{\mathbf{x}}_{k+1|k}$ stands for estimate of $\mathbf{x}_{k+1}$ given $U^{[k+1]}$ and $Z^{[k]}$, we first predict $\hat{\mathbf{x}}_{k+1|k}$ and then, when $\mathbf{z}_{k+1}$ is available, compute $\{\hat{\mathbf{x}}_{k+1|k+1}, \hat{\mathbf{x}}_{k|k+1}, \ldots, \hat{\mathbf{x}}_{k-q+2|k+1}\}$ as predict-correct frameworks. The ability of predict-correct frameworks to "go back" and compute an improved estimate $\hat{\mathbf{x}}_{k-q+2|k+1}$ than $\hat{\mathbf{x}}_{k-q+2|k-q+1}$ computed earlier offers the possibility that occurrence of an easier to detect event at instant $k+1$ can be used to learn correct mapping when $\mathbf{z}_{k-q+2}$ were either not available or were not informative.

Metrics called specificity ($SPC$), sensitivity ($SEN$) and Average Detection Rate ($ADR$) are used. Sensitivity is defined as $SEN \equiv Y^+ / (Y^+ + N^-)$ where $Y^+$ is the count of true positives and $N^-$ is the count of false negatives. Specificity is defined as $SPC \equiv N^+ / (N^+ + Y^-)$ where $N^+$ is the count of true negatives and $Y^-$ is the count of false positives. Finally $ADR \equiv (SPC + SEN)/2$.

The problem is one of computing parameters $\mathbf{w} = \{w_i\}_{i=1}^{i=M}$ of a parameterized function $f_\mathbf{w}$ such that when it is applied to the training input data $\mathbf{U} = \{\mathbf{u}_j\}_{j=1}^{j=N}$ some metric computed from values $\hat{\mathbf{Z}} = \{\hat{z}_j = f_\mathbf{w}(\mathbf{x}_j)\}_{j=1}^{j=N}$ returned by the function and supplied output values $\mathbf{Z} = \{z_j\}_{j=1}^{j=N}$ is optimized.

---

[2] Another application area with this characteristic is Deep Space Exploration. Once a explorer is launched with certain, preferably learning, computation engines, these computation engines have to learn to solve the problem even if the characteristics of the problems evolve as explorer travels deeper and deeper into outer space.

For example, it may be desired that sum of squared errors is minimized. Then the problem is, find $\hat{\mathbf{w}}$ where

$$\hat{\mathbf{w}} = \underset{\mathbf{w}}{\mathrm{argmin}} \sum \left(f_{\mathbf{w}}(\mathbf{u}_j) - z_j\right)^2 \quad (3)$$

Having computed $\hat{\mathbf{w}} = \{\hat{w}_i\}_{i=1}^{i=M}$, prediction is made as $\hat{z}_j = f_{\hat{\mathbf{w}}}(\mathbf{u}_j)$.

For the problem under consideration, $N >> M$, and a problem encountered is that of over-fitting wherein a $\hat{w}$ is determined that well minimizes the sum of squared errors for the training data (by learning too well the characteristics present) but then performs poorly on other valid data (that may exhibit a broader set of characteristics). To counter this, a *regularization* term may be added to the minimization and the problem may be written as

$$\hat{\mathbf{w}} = \underset{\mathbf{w}}{\mathrm{argmin}} \sum \left(\left(f_{\mathbf{w}}(\mathbf{u}_j) - z_j\right)^2 + \lambda g(\mathbf{w})\right) \quad (4)$$

If it is assumed that the relation of output $z$ to corresponding input $\mathbf{u}$ is *linear*, i.e.

$$z_j = \sum_{i=1}^{m} u_{ij} w_i = \mathbf{u}_j^T \mathbf{w}_j, \quad (5)$$

$\mathbf{u}_j$ s are drawn from independent identical distributions ($\mathbf{u}_j$ does not depend on any $\mathbf{u}_i$, $i \neq j$ and $\mathbf{u}_j$ for all $j$, are drawn from the same distribution), components of $\mathbf{u}_j$ are not linearly correlated and lastly the regularization function simply takes the sum of square of individual elements, i.e. $g(\mathbf{w}) = \sum_{i=1}^{m} w_i^2 = \|\mathbf{w}\|_2^2$, the problem can be expressed as

$$\hat{\mathbf{w}} = \underset{\mathbf{w}}{\mathrm{argmin}} \sum \left(\left(f_{\mathbf{w}}(\mathbf{u}_j) - z_j\right)^2 + \lambda g(\mathbf{w})\right)$$

$$= \underset{\mathbf{w}}{\mathrm{argmin}} \left((\mathbf{Uw} - \mathbf{z})^T (\mathbf{Uw} - \mathbf{z}) + \lambda \mathbf{w}^T \mathbf{w}\right). \quad (6)$$

Here $\lambda$ is a trade-off constant that trades-off training error minimization for regularization.

As expressed above the problem is called ridge-regression and has a simple closed form solution

$$\hat{\mathbf{w}} = \left(\mathbf{U}^T \mathbf{U} + \lambda \mathbf{I}\right)^{-1} \mathbf{U}^T \mathbf{y}. \quad (7)$$

The above solution represents an off-line training method and, as introduced above has following assumptions:

- $z_j$ is linearly dependent on $\mathbf{u}_j$,
- $\mathbf{u}_j$ s are i.i.d.,
- components of $\mathbf{u}_j$ are not linearly correlated,
- given training pairs ($\mathbf{u}_j$, $z_j$) are noise free and
- objective and regularization are as in Ridge Regression.

We know each of these to not hold for Epileptic Seizure prediction. Adaptability and therefore online training is desirable in seizure prediction methods. Only a highly complex feature can hope to achieve linear relation to prediction output. Individual feature vectors in a sequence of feature vectors are not independent. Training data is usually significantly noisy. Objectives such as maximizing average detection rate are not maximizing linear combination of distances.

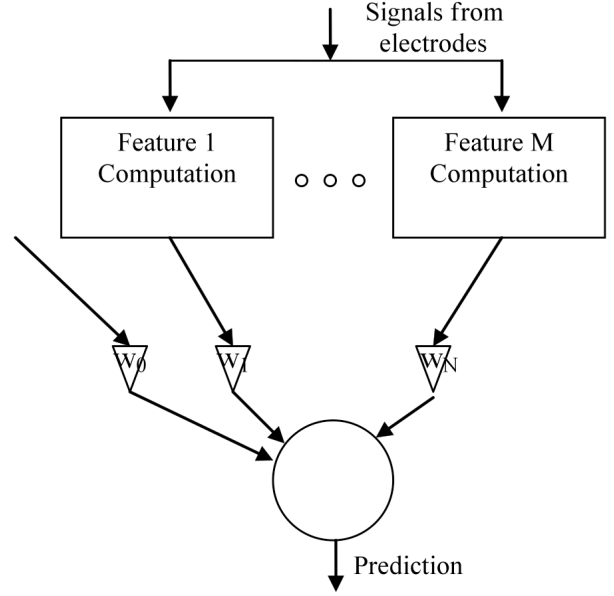

Figure 4: Prediction with manually engineered features

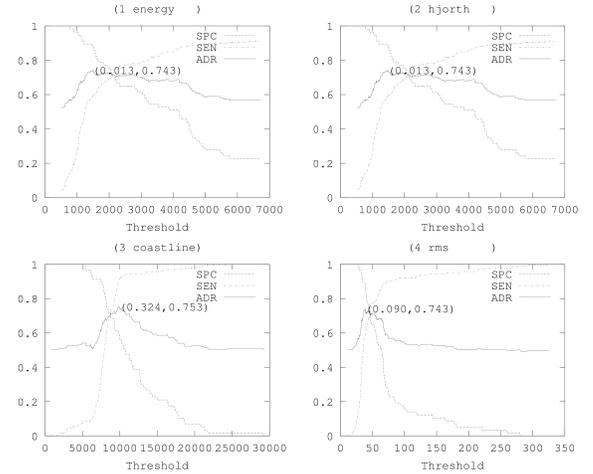

Figure 5: Metrics SPC, SEN and ADR versus illustrative manually engineered features on a sample of real data shows SPC, SEN and ADR with respect to four manually engineered features [17] – Energy, Hjorth Variance, Coastline and RMS Amplitude.

Complex objectives rarely have solution in closed form. As long as the objective is differentiable, we can employ a member of the Gradient Descent family of algorithms for solution. Convex functions, characterized by $f(ax_1 + bx_2) \leq af(x_1) + bf(x_2)$ are able to express a wider set of objectives. Carefully designed members of the Gradient Descent family can optimize convex objectives, can accommodate samples that are not independent [7] and even overcome noise and certain forms of non-convexity by taking advantage of the structure in the objective [15].

When components of $\mathbf{u}_j$ are not completely orthogonal, methods exist to orthogonalize them [13]. Most regression packages detect when some components are linearly correlated and either suggest components to eliminate or set respective coefficients to zero.

Ignoring for the moment that $\mathbf{u}_j$ s are not i.i.d. and employing manually engineered features, we can use organization shown in Figure 3 to perform prediction. The organization uses a single neuron. The weights of the neuron can be trained using a member of the Gradient Descent family of algorithms. This organization does not compute any further higher layer features beyond those manually engineered and makes predictions that are non-linear but not highly non-linear functions of the features.

Figure 5 shows SPC, SEN and ADR with respect to four manually engineered features 0 – Energy, Hjorth Variance, Coastline and RMS Amplitude. The non-smooth and gradient-discontinuous nature of the plot is from, among others, presence of noise and use of discrete 'counts' of 'events' – a 0 if did not occur and a 1 if it did - in the definition of the SPC and SEN.

When $\mathbf{u}_j$ s are not i.i.d., 'Time Series' methods consider that $\mathbf{u}_j$ is value at 'time instant' $j$, $\mathbf{u}_{j+1}$ is value at 'time instant' $j+1$ and so on. These methods take covariance explicitly into account.

Assumption of linear relation of $z_j$ to $\mathbf{u}_j$ can be eliminated by employing a Neural Network. Often inference requires computing an unknown and, possibly highly, non-linear function of some characteristics of the data. Neural Networks can efficiently find nonlinear mappings.

Feedback interconnections within Recurrent Neural Networks impart them the ability to retain information and enable them to perform inference tasks such as Epileptic Seizure Prediction involving time series such as Electroencephalographic data of brain activity.

Deep Neural Networks can automatically extract relevant high-level characteristics making it unnecessary to have to manually engineer characteristic extraction and avoiding loss of performance from characteristics being not matched to task at hand. However, most importantly, automatic feature extraction, addresses the need for adaptability whereby features suited to a patient are used. Furthermore, as the treatment progresses, the features used can also evolve to suit.

However both, the network being deep and network incorporating feedback, add to the difficulty of training the network. Approaches considered for training deep networks include combining and following unsupervised pre-training with supervised training [4]. The unsupervised pre-training tries to minimize energy as low-energy states correspond to parameters attaining values that capture structures present in data. The following training then is intended to further evolve the parameters to capture the desired input-output mapping. At core of both of these training algorithms is usually the backpropagation gradient descent.

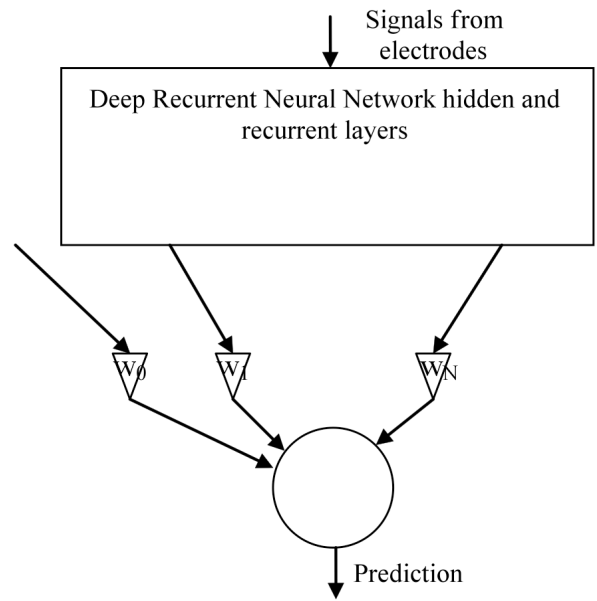

Figure 6: Automatic High-Level Feature Extraction Using Deep Recurrent Neural Network

Among the earliest methods considered for training Recurrent Networks is the Back-Propagation Through Time method [24] that can be considered the analogue of the Back Propagation method for Feed-Forward Networks. However the BPTT method had been believed to be prone to getting caught in many shallow local minima on the error hyper-surface in the parameter hyper-space. Traditionally, for local minima avoidance, unit activation and/or weight update are made stochastic and the activation or the weight are assigned, rather than the exact computed value, a value drawn from the applicable distribution with the computed values as the moments of the distribution.

It was first noted in late 80s that the Neural Networks are not unlike non-linear physical systems and that there exists a well-developed theory of estimating hidden state and estimating parameters of linear systems from noisy observations and somewhat less developed similar theories for nonlinear systems. A number of methods adhering to Predict-Correct framework (e.g. the celebrated Kalman Filter) have been developed over time that are optimal in the sense of Minimum Mean Squared Error for state only, parameter only and joint state and parameter estimation problems for linear systems. Optimal estimation is no longer computationally tractable when the system is nonlinear.

The realization that Recurrent Neural Networks are non-linear systems led to use of Extended Kalman Filter, popularly used for non-linear system identification, parameter estimation, etc for training Recurrent Networks. The Extended Kalman Filter has a set of well known limitations. The limitations stem from approximating the non-linear dynamics of systems as (locally) linear for purposes of deriving the moments of the posterior state distribution and from approximating the state distributions as Gaussian.

Time-series prediction has been addressed by Statistics, Dynamical Systems, Machine Learning, Signal Processing and Automatic Control communities. The need to address larger and more complex systems has led beyond Back-Propagation Through Time and first order term (in Taylor expansion) only Extended Kalman Filter [11] to including second order terms [14], avoiding explicit use of Jacobian and Hessian [14], Quadrature [1] and Cubature [1] forms, the Unscented [12] form, the marginalized Sigma Points [19] form, the Particle Bayesian [22] form, the Variational Bayesian [22] form and the Gaussian Process [3] form. While an improvement results, a conceptual and sometimes also a computational complexity cost has to be paid.

It is to be noted that greater the non-linearity greater are the errors introduced by the linear approximation. In case when process and measurement functions are deep neural networks, if the transformation from external inputs thru the collection of layers to final outputs is considered, the transformation represents a large degree of non-linearity.

Computing the derivatives is tedious and is considered another shortcoming of methods requiring derivatives.

It is needed to address both the eventual and the speed of convergence. The model of a system, that when supplied with some input with stationary statistics, is producing the same time varying output as the system, cannot be considered converged if one-step updates are non-zero.

### C. One-step and long-term time-series prediction

The problem comprises of being given an input time-series $\mathbf{U} = \{\mathbf{u}_i\}_{i=1}^{i=k}$ and noisy observations of corresponding output time-series $\mathbf{z} = \{z_i\}_{i=1}^{i=k}$ where observations may not be available at each instant of time $1 \leq i \leq k$ and the required result being $z_{k+q}, q \geq 1$. As an example, the input time-series may be intra-cranial electrode sensed electric potential, the output time-series values may only be available for certain $i$, may be whether the patient was experiencing an epileptic seizure[3] and the required result is whether the patient is about to experience an epileptic seizure $q$ instants from now.

### D. Extrapolating highly non-linear functions

Consider a function $y = f(x) = e^{a_1 e^{a_2 x + b_2} + b_1}$ specified as a sequence of functions: $y = f_1(y_1) = e^{a_1 y_1 + b_1}$ and $y_1 = f_2(x) = e^{a_2 x + b_2}$.

We are given value $y'$ of $y$ at $x = x'$ and it is desired to extrapolate value at $x'' = x' + \Delta x'$.

A linear approximation can be done and we can write $y'' = f(x'') = f(x' + \Delta x') = f(x') + (\partial f / \partial x)_{x = x'} \Delta x' + \cdots$.

It is easy to show that a more accurate approximation can

---

[3] Note that in an offline training situation, the EEG data may be examined by an expert encephalographer and intervals corresponding to seizure determined. In an online training context, the parameters may be initialized to values determined offline and predictions used only after stability is reached.

instead be obtained as a sequence of approximations

$y = f_1(y_1'') = f(y_1' + \Delta y_1') = f_1(y_1') + (\partial f_1 / \partial y_1)_{y_1 = y_1'} \Delta y_1' + \cdots,$

$y_1' = f_2(x'') = f_2(x' + \Delta x') = f_2(x') + (\partial f_2 / \partial x)_{x = x'} \Delta x' + \cdots,$

$\Delta y_1' = (\Delta x' / x') y_1'$, etc.

In the context of training multi-layer recurrent neural networks, where weight update is computed as $\mathbf{w} \leftarrow \mathbf{w} - \eta (\partial e / \partial \mathbf{w})$, this fact can be taken advantage of by including in the state, in addition to prior outputs of hidden layers, the current output of visible and hidden layers.

### E. High order Markov Processes

Given a process that receives input $\mathbf{u}_k$ at instant $k$ and produces output $z_k$, the process may employ internal state $\mathbf{x}_k$ to remember as much of past history as it needs to so that it can compute output $z_k$ considering only $\mathbf{x}_k$ and $z_k$. Expressed another way, $P(z_k | \mathbf{u}_k, \mathbf{u}_{k-1}, \cdots, \mathbf{u}_0) = P(z_k | \mathbf{x}_k, \mathbf{u}_k)$ and system output is considered independent of past input given the current internal state. Such processes are called first-order. However this may require, not unlike carefully manually engineering features, carefully manually engineering the internal state that system must remember. A solution is to require system to remember history going back $b$ instants. Then $P(z_k | \mathbf{u}_k, \mathbf{u}_{k-1}, \cdots, \mathbf{u}_0) = P(z_k | \mathbf{x}_k, \mathbf{u}_k, \mathbf{u}_{k-1}, \cdots, \mathbf{u}_{k-b+1})$ and the process is called of order $b$.

## III. DEEP RECURRENT NEURAL NETWORKS

Neural Networks comprise of one or more inputs, one or more neurons (referred to in this work as units), one or more outputs, directed or undirected interconnections among units, a weight associated with each undirected connection of or each directed connection into a unit (from an input or another unit) and each unit computing its activation by applying its activation function to weighted sum of inputs.

Feed-Forward Neural Networks are arranged in layers with units in a layer receiving inputs from only units in previous layer (or outside in case of the input layer). Feed-Forward networks process data presented at an instant independent of data presented at previous instants and so are unable to perform inference on time series data. In a Recurrent Network, some of the interconnections are from units in a layer through a delay to units in the same or an earlier layer.

As both stacking of recurrent layers and back-propagating have the effect of extracting information from input or from information extracted from input to consider alongside input, it is a question as to whether (i) either is more effective than the other and (ii) having both together adds to the approximation ability of the network.

Each unit in a layer associates a weight to and has as inputs a fixed bias, output of all units in the previous layer (external inputs in case of the visible layer) and, in case of the recurrent layer, its own outputs delayed by a time instant. Output of a unit $y_{x_i}$ is the result of applying a nonlinear function $g$ to the weighted sum of inputs $\text{in}_{ij}$ : $\text{out}_i = g\left(\sum w_{ij}\text{in}_{ij}\right)$.

The network has a visible layer, $N_L$ hidden layers and an output layer of units (neurons) with, respectively, $N_V$, $N_{H_l}$ $(1 \le l \le N_L)$ and $N_O$ units in the layer. The network receives inputs $y_{u_q}$, $1 \le q \le N_I$ and produces outputs $\hat{y}_{z_i}$, $1 \le i \le N_O$. Furthermore, $v_p$ is the $p$-th unit (or alternatively unit $p$) in the visible layer, $h_{l,j}$ is the unit $j$ in the hidden layer $l$, $z_i$ is the unit $i$ in the output layer, $u_q$ is the input $q$.

As a convention, $\mathbf{y}_x$ is the vector with elements $y_{x_i}$, $y_{x_i}$ is the output of unit $x_i$, $w_{x_1 x_2}$ is the weight from unit $x_2$ to unit $x_1$, $y_{h_{l,j}^{(-k)}}$ is the output at (past) instant $-k$ of the node $j$ in hidden layer $l$. Notionally, associated with the unit $j$ in hidden layer $l$ are $B$ memories that retain the previous outputs $y_{h_{l,j}^{(-B)}}$ through $y_{h_{l,j}^{(-1)}}$ of the unit. $B$ is referred to as the extent of backpropagation.

Inputs to each unit in any layer include a fixed bias 1. We use the same symbol as used to denote the output signal vector of a layer, but with a tilde accent, to denote a signal vector with 1 (corresponding to the fixed bias) as an additional and the first element and remaining elements same as the original signal vector. Then

$$\mathbf{y}_v = \mathbf{W}_{v,u}\tilde{\mathbf{y}}_u \tag{8}$$

$$\mathbf{y}_{h_{N_L}} = \mathbf{W}_{h_{N_L},v}\tilde{\mathbf{y}}_v + \mathbf{W}_{h_{N_L},h_{N_L}^{(-1)}}\mathbf{y}_{h_{N_L}^{(-1)}} \tag{9}$$

$$\mathbf{y}_{h_{N_L}^{(-1)}} = \mathbf{W}_{h_{N_L}^{(-1)},v^{(-1)}}\tilde{\mathbf{y}}_{u^{(-1)}} + \mathbf{W}_{h_{N_L}^{(-1)},h_{N_L}^{(-2)}}\mathbf{y}_{h_{N_L}^{(-2)}} \tag{10}$$

Reader may notice that we have chosen to qualify, rather than the symbol denoting the output signal of a node, the symbol denoting the node with the superscripts $(-b)$ denoting value of "the node at past instant $b$" rather than "value at past instant $b$" of the node. This is deliberate and is to suggest that we consider the network to have been unfolded back and there to exist $B$ copies, one corresponding to each of $B$ instants spanning the extent of the backpropagation, of the complete "sub-network" comprising of the visible and each of the hidden layers. As a result, the notation allows not only vectors $\mathbf{y}_{u^{(-1)}},\cdots,\mathbf{y}_{u^{(-B)}}$ to be distinguished from the vector $\mathbf{y}_u$, but also weight matrices $\mathbf{W}_{v^{(-1)},u^{(-1)}},\cdots,\mathbf{W}_{v^{(-B)},u^{(-B)}}$ to be distinguished from $\mathbf{W}_{v,u}$. If this distinction was not made (10) above will be

$$\mathbf{y}_{h_{N_L}^{(-1)}} = \mathbf{W}_{h_{N_L}^{(-1)},v^{(-1)}}\tilde{\mathbf{y}}_{v^{(-1)}} + \mathbf{W}_{h_{N_L}^{(-1)},h_{N_L}^{(-2)}}\mathbf{y}_{h_{N_L}^{(-2)}} \tag{11}$$

Outputs of all the units as well as the weights associated with the inputs to the units are considered collected into a joint-state-and-parameter-vector. This vector at instant $k$ is denoted as $\mathbf{x}^{(k)}$, estimate of the vector is denoted as $\hat{\mathbf{x}}^{(k)}$ and element $i$ of the same is denoted as $\left(\hat{\mathbf{x}}^{(k)}\right)_i$. Final outputs of the network are also considered collected into a vector. This vector at instant $k$ is denoted as $\hat{\mathbf{z}}_k$.

A significant component of any method making use of the derivatives are the derivative computations

$$\left(\mathbf{F}^{(k)}\right)_{ij} = \frac{\partial\left(\hat{\mathbf{x}}^{(k)}\right)_i}{\partial\left(\hat{\mathbf{x}}^{(k)}\right)_j} \tag{12}$$

$$\left(\mathbf{H}^{(k)}\right)_{ij} = \frac{\partial\left(\hat{\mathbf{z}}^{(k)}\right)_i}{\partial\left(\hat{\mathbf{x}}^{(k)}\right)_j} \tag{13}$$

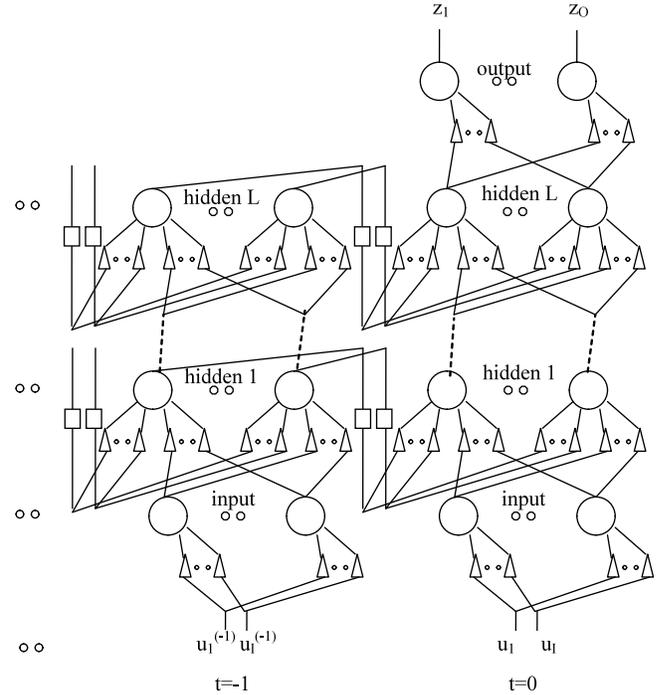

Figure 7: Back-Propagation Through Time and Space in Deep Recurrent Neural Networks

Visualization of dependencies considered in the derivative computation is eased by considering the nodes making up the network to be arranged as columns of stacks of layers of linear array of nodes (Figure 7). The input layer is at the bottom, the hidden layers in the middle and the output layer at the top. The rightmost column corresponds to current instant and the columns to left to successively previous instants. A notion of distance can be associated with any pair of units and of a unit and a weight. A unit in the (present-instant, output-layer) is at a distance of 1 from any unit in the (present-instant, hidden layer), at a distance 2 from any unit in (present instant, visible layer), also at a distance 2 from (previous instant, hidden layer) and so on. Similarly the instant $-1$ weight $w_{h_{N_L,i}^{(-1)},v_p^{(-1)}}$ is at a distance 1 from (instant $-1$, hidden layer $N_L$) unit $j$, distance 2 from any unit in the (present instant, hidden layer $N_L$), also distance 2 from any unit in the (instant $-1$, layer $N_L - 1$), distance 3 from any unit in the (present instant, layer $N_L - 1$), also distance 3 from any unit in (instant $-1$, layer $N_L - 2$), etc.

Output at instant $k$ depends on parameters (weights) and state at $k$. The state at instant $k$ in turn depends on weights and state at $k-1$. Once weights have been updated at instant $k$, a new output at $k$ can be computed. To compute this output, the state at $k-1$ is required. However the state at $k-1$ itself can be recomputed using the updated weights and the state at $k-2$ and so on.

Given the error in output at $k$, part of the error can be attributed to errors in weights and part to error in the (estimated) state at $k$. The error in state at $k$ in turn depends in part on errors in weights and part on error in state at $k-1$.

If the mapping from input to output is invariant, the weights at instants $k-1$ and $k-2$ are the same. However if the mapping itself is evolving then the two sets of weights are not the same.

Once the weights corresponding to present and to each past instant are made different and the present error is computed, weights at instant $k-1$ can be overwritten with weights at instant $k$ and present instant weights can be updated to next instant weights. Alternatively, the back-propagation itself can be used to compute weights at present as well as each retained past instant.

The mapping $\mathbf{z}^{(k)} = h(\mathbf{x}^{(k-1)}, \mathbf{u}^{(k)})$ may be time-invariant, may need to be learnt and, once learnt, does not change. Alternatively the mapping may continuously evolve.

Computing only next instant weights by updating present instant weights and overwriting instant $k-1$ weights with instant $k$ weights is consistent with models that allow the mapping to be learnt and then assume them to remain constant. Computing all the weights using back-propagation is consistent with models where mapping may continuously evolve.

### A. Universality of single hidden layer feed-forward network, bounds on approximation error and multilayer networks

Single hidden layer feed-forward networks using the monotone cosine squashers are capable of embedding as a special case a Fourier network that yields a Fourier series approximation to a given function at its outputs [8] and thus possesses all the approximation properties of Fourier series representations.

A feed-forward network with a single hidden layer of continuous sigmoidal nonlinear squashers (in addition to linear visible and output layers) are complete in the space of functions on $R^d$ that have continuous and uniformly formed derivatives up through order $m$ [10].

For a network with a hidden layer with $n$ units, the approximation error is bounded by $c_f'/n$ [2] where $c_f'$ depends on the norm of the Fourier Transform of the function being approximated

$$c_f' = (2rC_f)^2 = \left(2r\int |\omega| \|\tilde{F}(d\omega)\|\right)^2 \tag{14}$$

where $\tilde{F}(\mathbf{w})$ Fourier distribution of function $f(\mathbf{x})$, $\mathbf{x} \in R^d$, such that $f(\mathbf{x}) = \int e^{i\mathbf{wx}} \tilde{F}(d\mathbf{w})$, $\tilde{F}(d\mathbf{w}) = e^{i\theta(\mathbf{w})} F(d\mathbf{w})$, $F(\omega)$ is the magnitude distribution, $\theta(\omega)$ the phase at frequency $\omega$ and $\mathbf{w}$ the vector of frequencies $(\omega_1, \cdots, \omega_d)$.

Let $\Gamma$ be the class of functions $f$ on $R^d$ for which $C_f = \int |\mathbf{w}| \|\tilde{F}(d\mathbf{w})\| < \infty$ and $\Gamma_C$ the class of functions $f$ on $R^d$ for which $C_f = \int |\mathbf{w}| \|\tilde{F}(d\mathbf{w})\| < C$.

Then, for every choice of fixed basis functions $h_1, h_2, \cdots, h_n$ [2]

$$\sup_{f \in \Gamma_C} s(f, \text{span}(h_1, h_2, \cdots, h_n)) \geq \kappa \frac{C}{d} \left(\frac{1}{n}\right)^{\left(\frac{1}{d}\right)} \tag{15}$$

where

$$s(f, G) = \inf_{g \in G} s(f, g) \tag{16}$$

$$s(f, g) = \int (f(x) - g(x))^2 dx \tag{17}$$

Let,

$B$ be a bounded set in $R^d$ which contains $x = 0$,

$\Gamma_B$ the set of functions $f$ on $R^d$ for which $f(x) = f(0) + \int (e^{i\mathbf{w}\cdot\mathbf{x}} - 1) \tilde{F}(d\mathbf{w})$ holds for $x \in B$ and some $\tilde{F}(d\mathbf{w}) = e^{i\theta(\mathbf{w})} F(d\mathbf{w})$,

$\Gamma_{C,B}$ the set of functions $f$ on $R^d$ for which

$$\int |\omega|_B F(d\omega) \leq C > 0 \tag{18}$$

where

$$|\mathbf{w}|_B = \sup_{\mathbf{x} \in B} \{\mathbf{w}\cdot\mathbf{x}\}, \tag{19}$$

then for every function $f \in \Gamma_{C,B}$, for every probability measure $\mu$, for every sigmoidal function $\phi$, for every $n > 0$, there exists a linear combination of functions $f_n$ of the form

$$f_n(\mathbf{x}) = \sum_{k=1}^{n} c_k \phi(\mathbf{a}_k \mathbf{x} + b_k) + c_0$$

such that [2]

$$\int (f_n(\mathbf{x}) - f(\mathbf{x}))^2 \mu(d\mathbf{x}) < c_f'/n = (2rC_f)^2/n \tag{20}$$

where $B_r = \{\mathbf{x} \in R^d \| |\mathbf{x}| \leq r\}$.

Hence adding a sigmoidal layer improves the rate from $(1/n)^{1/d}$ to $1/n$. Will adding another layer improve further? By how much?

In a one linear layer network we take a weighted sum of inputs and add in a bias $y_{n0} = f_{n0}(\mathbf{x}) = \mathbf{a}^T \mathbf{x} + a_0$ where for ease of explanation we number the elements of $\mathbf{a}$ and $\mathbf{x}$ starting at $1$.

Equation (7) bounds below the approximation error of one linear layer network.

In a 1 hidden layer network, we take $n$ distinct copies of one linear layer networks, feed the output of each into a nonlinearity, take a weighted sum of the outputs of the nonlinearities and add in a bias

$$f_{n^1}(\mathbf{x}) = \mathbf{b}^T \cdot \left(\phi(\mathbf{a}_1^T \mathbf{x} + a_{10}), \cdots, \phi(\mathbf{a}_n^T \mathbf{x} + a_{n0})\right)^T + b_0$$
$$= \mathbf{b}^T \cdot \left((f_{n^0,1}(\mathbf{x})), \cdots, (f_{n^0,n}(\mathbf{x}))\right)^T + b_0. \quad (21)$$

Equation (8) bounds below the rate of error. Note that here we are concerned with inherent representability using multi-layer neural networks, are not concerned with error in determination of weights and assume that values of weights minimizing error, if they exist, can be found.

Consider $\int (f_{n^2}(\mathbf{x}) - f(\mathbf{x}))^2 \mu(d\mathbf{x})$. $f_{n^2}(\mathbf{x})$ is the function computed by the two hidden layers network. The two hidden layers network has (logically) $n$ instances of the one hidden layer network as sub-networks. Let $f_{n^2}^*$ be the function computed by the two hidden layers network when it achieves minimum error and $(\tilde{f}_{n^1,1}(\mathbf{x}), \cdots, \tilde{f}_{n^1,n}(\mathbf{x}))$ be the functions needed to be computed by its one hidden layer sub-networks when the two hidden layers network achieves minimum error. Let $(f_{n^1,1}(\mathbf{x}), \cdots, f_{n^1,n}(\mathbf{x}))$ be the actual functions computed by the one hidden layer sub-networks. Similarly let $(\tilde{f}_{n^0,1}(\mathbf{x}), \cdots, \tilde{f}_{n^0,n}(\mathbf{x}))$ be the functions needed to be computed by their one linear layer sub-networks when the one hidden layer network achieves minimum error relative to $(\tilde{f}_{n^1,1}(\mathbf{x}), \cdots, \tilde{f}_{n^1,n}(\mathbf{x}))$ and $(f_{n^0,1}(\mathbf{x}), \cdots, f_{n^0,n}(\mathbf{x}))$ be the actual functions computed by the one linear layer sub-networks.

Let us denote
$$\psi_{n^i}(\mathbf{x}) = \left(\phi(f_{n^i,1}(\mathbf{x})), \cdots, \phi(f_{n^i,n}(\mathbf{x}))\right)$$
$$\tilde{\psi}_{n^i}(\mathbf{x}) = \left(\phi(\tilde{f}_{n^i,1}(\mathbf{x})), \cdots, \phi(\tilde{f}_{n^i,n}(\mathbf{x}))\right)$$
$$\psi_{n^i}^+(\mathbf{x}) = \left(\phi(f_{n^i,1}^+(\mathbf{x})), \cdots, \phi(f_{n^i,n}^+(\mathbf{x}))\right)$$
$$\psi_{n^i}^-(\mathbf{x}) = \left(\phi(f_{n^i,1}^-(\mathbf{x})), \cdots, \phi(f_{n^i,n}^-(\mathbf{x}))\right)$$

Then
$$f_{n^2}^*(\mathbf{x}) = \tilde{\psi}_{n^1}(\mathbf{x}) \cdot \mathbf{c} + c_0, \quad (22)$$
$$f_{n^1}(\mathbf{x}) = \psi_{n^0}(\mathbf{x}) \cdot \mathbf{b} + b_0, \quad (23)$$
$$\int (f_{n^1,k}(\mathbf{x}) - \tilde{f}_{n^1,k}(\mathbf{x}))^2 \mu(d\mathbf{x}) < c_f'/n^d = (2rC_f)^2/n^d \quad (24)$$
and
$$\sup_{f \in \Gamma_C} s(\tilde{f}_{n^0}, f_{n^0}) \geq \kappa \frac{C}{d} \left(\frac{1}{n}\right)^{\left(\frac{1}{d}\right)}. \quad (25)$$

Hence in spite of the one linear layer subnetworks having error bounded below by $\kappa(C/d)(1/n)^{(1/d)}$, 1 hidden layer subnetworks manage to achieve error $\int (f_{n^2}(\mathbf{x}) - f(x))^2 \phi(d\mathbf{x})$ bounded above by $(2rC_f)^2/n$.

What if the errors of functions computed by the subnetworks are smaller e.g. $(2rC_f)^2/n$ versus $\kappa(C/d)(1/n)^{(1/d)}$?

Note function computed is weighted sum of $n$ terms
$$f_{n^1}(\mathbf{x}) = \psi_{n^0}(\mathbf{x}) \cdot \mathbf{b} + b_0 \quad (26)$$

For squared minimum error
$$(f_{n^1}^*(\mathbf{x}) - f(x))^2 = \inf_{f_{n^0,1}, \cdots, f_{n^0,n}, \mathbf{b}, b_0} (\psi_{n^0}(\mathbf{x}) \cdot \mathbf{b} + b_0 - f(\mathbf{x}))^2$$
$$= (\tilde{\psi}_{n^0}(\mathbf{x}) \cdot \mathbf{b} + b_0 - f(\mathbf{x}))^2 \quad (27)$$

While there can be cases of small2 input errors and errors adding, such that a large output error results. However in expectation and averaged over large number of points, smaller input (sub-network) errors will result in small output (network) errors.

If
$$\int (\tilde{\psi}_{n^0}(\mathbf{x}) \cdot \mathbf{b} + b_0 - f(\mathbf{x}))^2 d\mu(x)$$
$$= \inf_{f_{n^0,1}, \cdots, f_{n^0,n}, \mathbf{b}', b_0'} \int (\psi_{n^0}(\mathbf{x}) \cdot \mathbf{b}' + b_0' - f(\mathbf{x}))^2 d\mu(x) \quad (28)$$
and
$$\int (f_{n^0,k}^-(\mathbf{x}) - \tilde{f}_{n^0,k}(\mathbf{x}))^2 \mu(d\mathbf{x})$$
$$< \int (f_{n^0,k}^+(\mathbf{x}) - \tilde{f}_{n^0,k}(\mathbf{x}))^2 \mu(d\mathbf{x}) \quad (29)$$
then
$$\int (\psi_{n^0}^-(\mathbf{x}) \cdot \mathbf{b} + b_0 - f(\mathbf{x}))^2 d\mu(x)$$
$$< \int (\psi_{n^0}^+(\mathbf{x}) \cdot \mathbf{b} + b_0 - f(\mathbf{x}))^2 d\mu(x) \quad (30)$$
Consider
$$(\psi_{n^0}^+(\mathbf{x}) \cdot \mathbf{b} + b_0 - f(\mathbf{x}))^2 - (\psi_{n^0}^-(\mathbf{x}) \cdot \mathbf{b} + b_0 - f(\mathbf{x}))^2$$
$$= (\psi_{n^0}^+(\mathbf{x}) \cdot \mathbf{b} + b_0 - f(\mathbf{x}))^2 - (\tilde{\psi}_{n^0}(\mathbf{x}) \cdot \mathbf{b} + b_0 - f(\mathbf{x}))^2$$
$$- (\psi_{n^0}^-(\mathbf{x}) \cdot \mathbf{b} + b_0 - f(\mathbf{x}))^2 + (\tilde{\psi}_{n^0}(\mathbf{x}) \cdot \mathbf{b} + b_0 - f(\mathbf{x}))^2 \quad (31)$$
$$= ((\psi_{n^0}^+(\mathbf{x}) + \tilde{\psi}_{n^0}(\mathbf{x})) \cdot \mathbf{b} + 2b_0 - 2f(\mathbf{x})) \times ((\psi_{n^0}^+(\mathbf{x}) - \tilde{\psi}_{n^0}(\mathbf{x})) \cdot \mathbf{b})$$
$$- ((\psi_{n^0}^-(\mathbf{x}) + \tilde{\psi}_{n^0}(\mathbf{x})) \cdot \mathbf{b} + 2b_0 - 2f(\mathbf{x})) \times ((\psi_{n^0}^-(\mathbf{x}) - \tilde{\psi}_{n^0}(\mathbf{x})) \cdot \mathbf{b})$$
$$> 0 \quad (32)$$

The last step follows from each (inner) term of both factors of the first (outer) term of the above expression being greater than or equal to the corresponding (inner) term of the second (outer) term.

Hence (11) follows.

Intuitively difference between
$$\int \text{span}\left(\phi(f_{n^{L-1},1}(\mathbf{x})), \cdots, \phi(f_{n^{L-1},n}(\mathbf{x}))\right) d\mu(\mathbf{x})$$
and
$$\int \text{span}\left(f_{n^{L-1},1}(\mathbf{x}), \cdots, f_{n^{L-1},n}(\mathbf{x})\right) d\mu(\mathbf{x})$$ over the entire

hypersphere $(0,1)^d$ can be understood by considering a function $f(\mathbf{x})$ of $\mathbf{x}$ which remains equal to some value with large magnitude $G$ nearly everywhere but dips steeply down to 0 at some number of (related to $n$) points. A linear function can only descend and then ascend back up at rate determined by the linear coefficients and will produce large error. Note nonlinearity such as $\phi(\mathbf{x}) = \tanh(\mathbf{x})$ is not convex. However $d\phi(\mathbf{x})/d\mathbf{x}$ is. Furthermore, nonlinearity may still be convex/concave between $-(n-2)\pi < \mathbf{x} < -(n-1)\pi$ and $-(n-1)\pi < \mathbf{x} < -n\pi$.

IV. BACK-PROPAGATION THRU TIME AND SPACE

For Back-Propagation Thru Time and Space Gradient Descent Training, weight $w_{ij}$ is updated according to $w_{ij} \leftarrow w_{ij} - \eta(\partial e / \partial w_{ij})$ where $e = \hat{z} - z$ is the error, $\hat{z}$ is the output of the single output network at the instant when input $\mathbf{u}$ has been applied at its input and $(\mathbf{u}, z)$ is a training input-output pair.

Visualization of dependencies considered in the derivative computation is eased by considering the nodes making up the neural network to be arranged as columns of stacks of layers of linear array of nodes (Figure 7). The output layer is at the top, the hidden layers in the middle and the input layer at the bottom. The rightmost column corresponds to current time and the columns to left to successively previous times.

A notion of distance can be associated with any pair of units and any pair of a unit and a weight. A unit in the (present-instant, output-layer) is at a distance of 1 from any unit in the (present-instant, hidden layer $N_L$), at a distance 2 from any unit in (present instant, visible layer), also at a distance 2 from (previous instant, hidden layer $N_L$) and so on. Similarly the present instant weight $w_{v_p u_q}$ is at a distance 1 from (present instant, visible layer) unit $p$, distance 2 from any unit in the (present instant, hidden layer $N_L$), distance 3 from any unit in the (present instant, hidden layer $N_L - 1$), distance 3 from any unit in (previous instant, hidden layer $N_L$), etc. Hence the distances here are partially over space and partially over time.

Details of computation of the derivatives are in the Appendix. These suggest a dynamic programming method for computing the derivatives comprising of traversing the graph from inputs to outputs, computing derivative of unit outputs wrt distance 1 away unit outputs, distance 1 away weights and distance 2 away weights. Derivatives wrt unit outputs at distance greater than 1 away and weights at distance greater than 2 away are computed by taking weighted sums over derivatives of distance 1 away unit outputs.

A. *Adaptation and Autonomous Self Improvement*

BPTT implies that at instant $k$, we not only have $W_{H,i,k}$ (vector of weights for the hidden layer unit $i$ at instant $k$) but also $W_{H,i,k-1}, \ldots, W_{H,i,k-q+1}$ and we compute $y_{H,i,k-q+1}, \ldots, y_{H,i,k-1}$, outputs of hidden layer unit $i$ at instants $k - q + 1$ through $k - 1$. When the weights are updated, not only $W_{H,i,k}$ but also $W_{H,i,k-1}, \ldots, W_{H,i,k-q+1}$ are updated and $y_{H,i,k-q+1}, \ldots, y_{H,i,k+1}$ are recomputed. $y_{H,i,k+1}$ and $\hat{\mathbf{z}}_k$, to be computed at instant $k + 1$ now make use of recomputed weights and history.

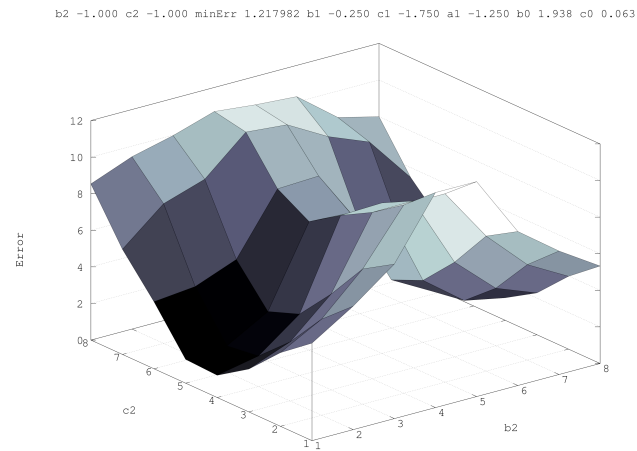

Figure 8: Minimum error (over all output layer weights) versus hidden layer weights for input weights (-1.5, -1.5)

The ability of predict-correct frameworks to "go back" and compute an improved estimate $\hat{\mathbf{x}}_{k-q+2|k+1}$ than the $\hat{\mathbf{x}}_{k-q+2|k-q+1}$ offers the possibility that occurrence of an easier to detect event at instant $k + 1$ can be used to learn a mapping that produces a better $\hat{\mathbf{z}}_{k-q+2}$ than when $\mathbf{z}_{k-q+1}$ alone was available. This is especially relevant to seizure onset prediction as determining that a seizure is about to occur is harder for both humans and computers than determining that a seizure is in progress.

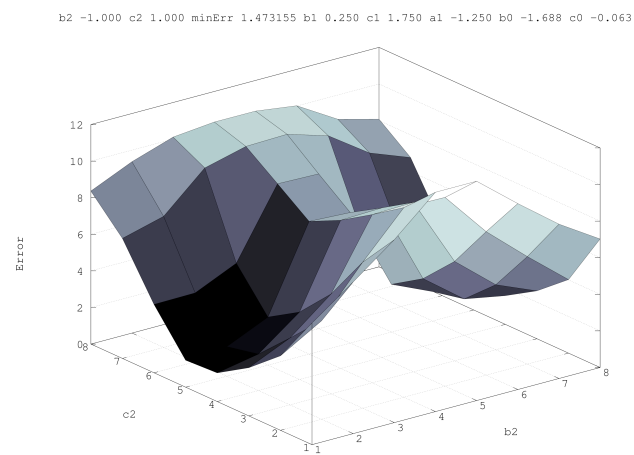

Figure 9: Minimum error (over all output layer weights) versus hidden layer weights for input weights (-1.5, 1.5)

## V. EXPERIMENTAL RESULTS

The first set of experiments explored the geometry of the parameter space first with (hidden layers 1, back-propagation extent 1), (hidden layers 1, back-propagation extent 2) and finally with (hidden layers 2, back-propagation extent 1). The visible, hidden and output layers each had one unit. Units in visible and output layers had one input from outside or the previous layer and one bias input. The unit in the hidden layer had one input from the visible layer, one bias input and one input that was its own output delayed by one instant. There was one input to and one output from the network. Hence there are in total 7 weights in the first, 12 in the second and 10 in the last case. Each weight corresponded to one dimension. A point in the first parameter space was specified by 7 coordinates. Each coordinate was restricted to one of a set of discrete values. Range spanned by all of the coordinates was restricted to a subrange of [-2, 2). For different coordinates, different number of discrete values evenly distributed across the range were visited during the systematic exploration.

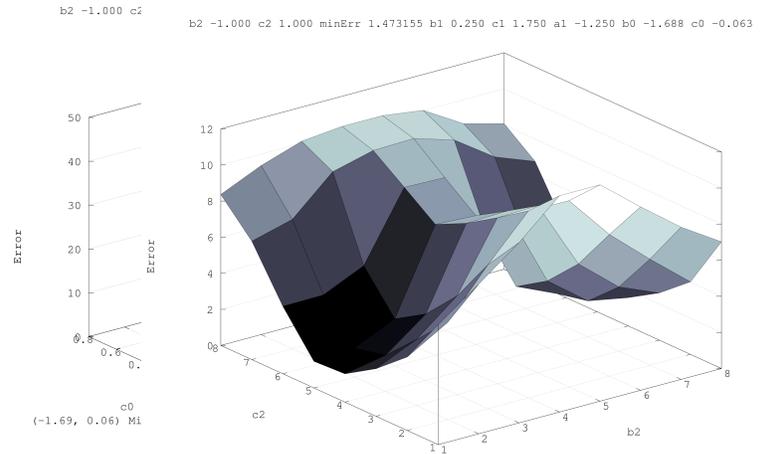

Figure 12: Err Figure 11: Predicted versus noise corrupted actual weights corresp

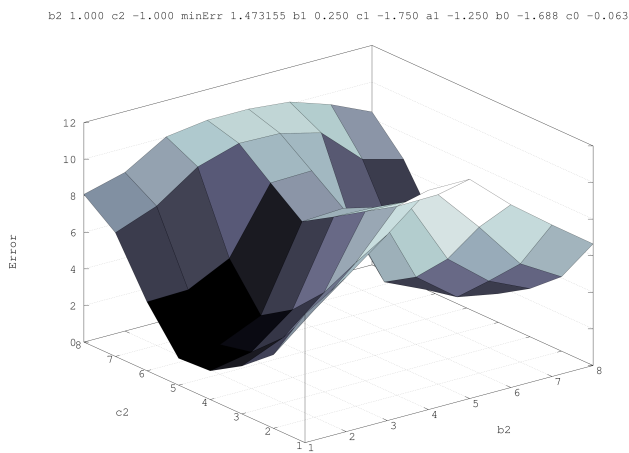

Figure 10: Minimum error (over all output layer weights) versus hidden layer weights for input weights (1.5, -1.5)

For the first two coordinates, corresponding to the weights associated with the visible layer unit, typically a small number (e.g. 2), for the next 3 coordinates corresponding to the hidden layer, a medium number (e.g. 8) and finally for the last two coordinates corresponding to the output layer a larger (e.g 32) discrete values were visited. Example number of values mentioned above resulted in $2^2 \times 8^3 \times 32^2 = 2^{18}$ different points in the parameter space to be visited.

For each of the 64 vectors of values of two of the three hidden layers weights – corresponding to the recurrent input and input from visible layer - minimum error was determined separately for each of 4 values of the weight pair corresponding to weights associated with the visible layer unit. These are plotted in Figures 8 through 11.

Next with the vector of visible and the hidden layer weights set to the vector corresponding to minimum error determined above, the error at the 1024 output layer weight pairs is plotted in Figure 13.

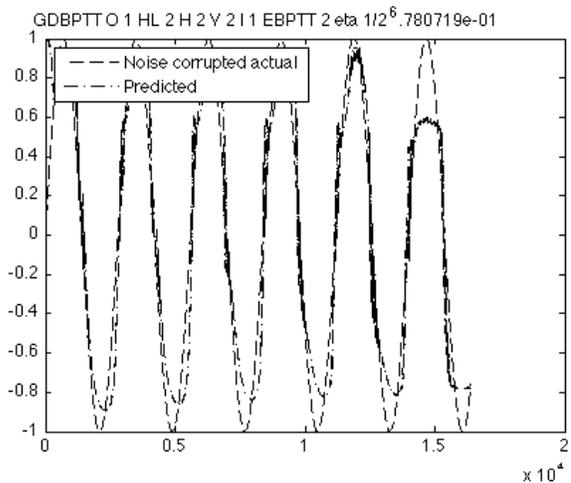

Figure 13: Predicted versus noise corrupted actual

Figures 8 through 11 have a bearing on the question as to whether with visible and hidden layer weights initialized randomly and output layer weights learnt using gradient-descent-backpropagation-through-time suffices to minimize error to arbitrarily small values. These present empirical evidence that it is not so. Arbitrary initialization of 'lower' layers and only learning weights of higher layers lower bounds the error, such that smaller error cannot be achieved.

Figure 12 has a bearing on the question whether with 'lower' layer weights set correctly, if the values of the error corresponding to different values of output layer weights are considered to lie on a manifold, what does this manifold look like. As the output layer comprises of sigmoidal units as well, the manifold is non-convex.

Simulation study was conducted with synthetic and real data. The synthetic data was the Sine and Mackey Glass function generated series. In case of the sine function generated series, the run was conducted upto $T = 8192$. However during $7T/8 \le k \le T$, prediction was performed but measurement update was not performed. We see that (Figure 11) predictions match the true data well. Even in case of the Mackey Glass function generated series, which is not simply periodic, predictions match the true data (Figure 14).

At each instant, input data is received, output is estimated and, if actual output is made available, the parameters are updated. It can be argued that with input with stationary statistics, if parameters do not attain stable values, correct output being estimated does not mean that input to output mapping has been learnt. As the plot of evolution of weights (Figure 14) shows, stable values are attained and the input-output mapping is truly being learnt.

The second set of simulation experiments are with real Electroencephalogram data collected at the Freiburg university [26]. There are 6 channels in all the recordings because of use of 6 electrodes. Each recording has been examined by an expert encephalographer and occurrences of seizure identified. Each recording covers a nearly continuous length of time that included several Ictal as well as inter-Ictal intervals. To shorten the run-times, middle of the inter-seizure intervals were shortened (e.g. to a unit variance Gaussian random number with a mean 262144 samples rather than original intervals exceeding 921600 samples) but intervals leading upto and the seizure intervals were included unmodified. For purposes of the simulation experiment, for each recording, measurement updates were performed through first several seizures ("training" seizures), measurement updates were stopped sometime before the occurrence of last few seizures ("test" seizures) and seizure prediction was compared to determination made by the expert encephalographer. Table presents summary of these experiments.

These set of experiments were repeated for prediction ahead into future by different amounts of time. The three sets presented in the table corresponded to prediction ahead by 1, 1025 and 2049 instants. High ADRs were achieved for each of these cases. To study whether, in spite of quantitatively similar performance for prediction ahead into future by different amounts of time, whether performance differed qualitatively. Figure 16, Figure 17 and Figure 18 show close-up of when in network determination recordings transition from no seizure to seizure when predicting ahead by 1, 1025 and 2049 instants.

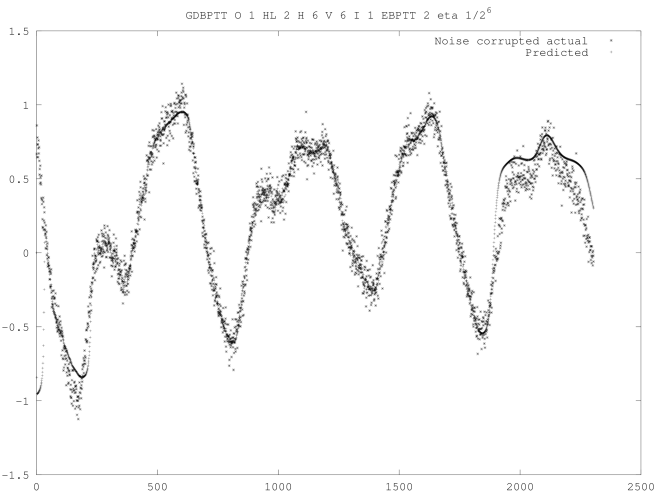

Figure 14: Predicted versus noise corrupted actual data for the Mackey Glass series

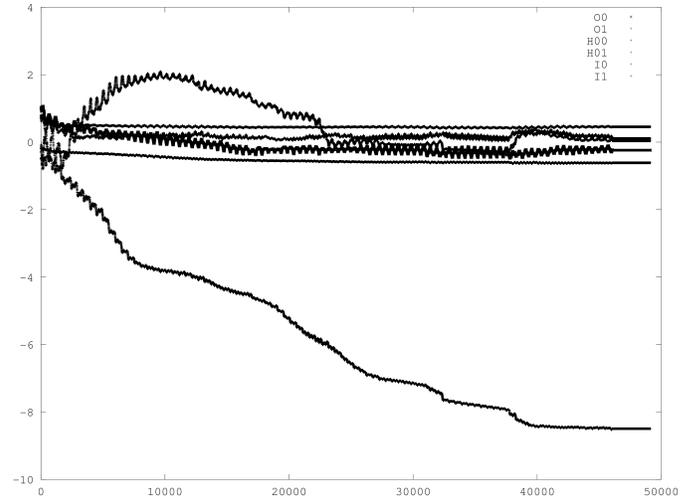

Figure 15: Evolution of weights

TABLE 1: SIMULATION EXPERIMENTS WITH EPILEPTIC SEIZURE DATA

| EXPERIMENT | RECORD NAME | RECORD LENGTH | PREDICT AHEAD | TRAIN SAMPLES | TEST SAMPLES |
|---|---|---|---|---|---|
| 1 | Pat003 | 2361858 | 1 | 2128306 | 233552 |
| 2 | Pat003 | 2361858 | 1025 | 2128306 | 233552 |
| 3 | Pat003 | 2302384 | 2049 | 2128304 | 174080 |

TABLE 2
REAL EPILEPTIC SEIZURE DATA TRAINING SEGMENTS

| EXPERIMENT | TRAIN SEIZURES | TRAIN SAMPLES MARKED SEIZURE |
|---|---|---|
| 1 | 3 | 111560 |
| 2 | 3 | 111560 |
| 3 | 3 | 111560 |

TABLE 3
REAL EPILEPTIC SEIZURE DATA TEST RESULTS

| EXPERIMENT | TRUE POSITIVE, FALSE NEGATIVE | SPC | TRUE NEGATIVE, FALSE POSITIVE | SEN | ADR |
|---|---|---|---|---|---|
| 1 | 6224, 175 | 0.97 | 227010, 143 | 1 | 0.986011 |
| 2 | 6308, 91 | 0.97 | 227106, 47 | 1 | 0.992786 |
| 3 | 6224175 | 0.97 | 167538, 143 | 1 | 0.99 |

The blue curve corresponds to expert determination and the green curve to network determination. Similarly Figure 19, Figure 20 and Figure 21 show close-up of when in network determination recordings transition from seizure to no seizure when predicting ahead by 1, 1025 and 2049 instants.

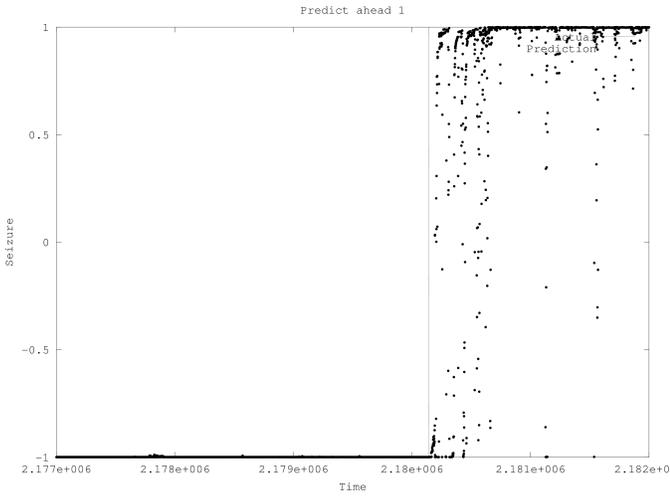

Figure 16: 1 instant ahead prediction of seizure onset – predicted (green) versus expert determination (blue)

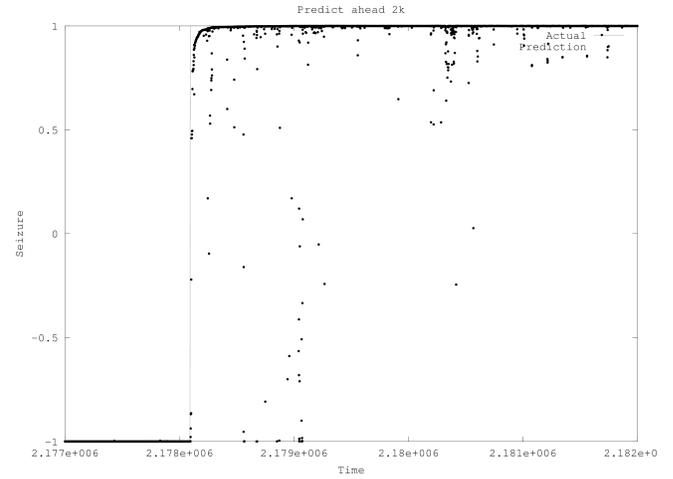

Figure 18: 2048 instants ahead prediction of seizure onset – predicted (dots) versus expert determination (solid line)

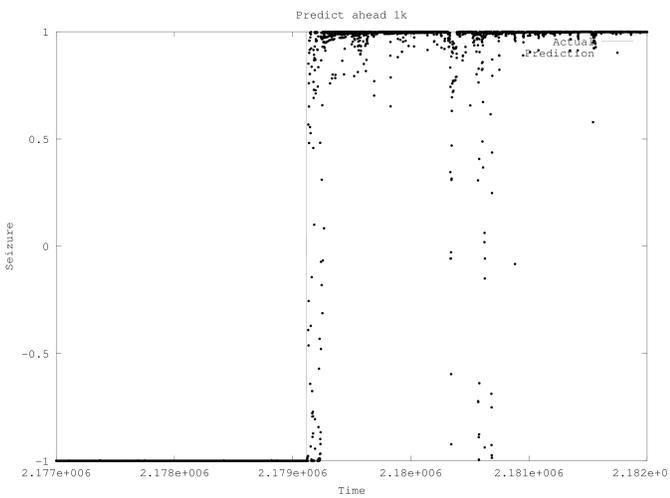

Figure 17: 1024 instants ahead prediction of seizure onset – predicted (green) versus expert determination (blue)

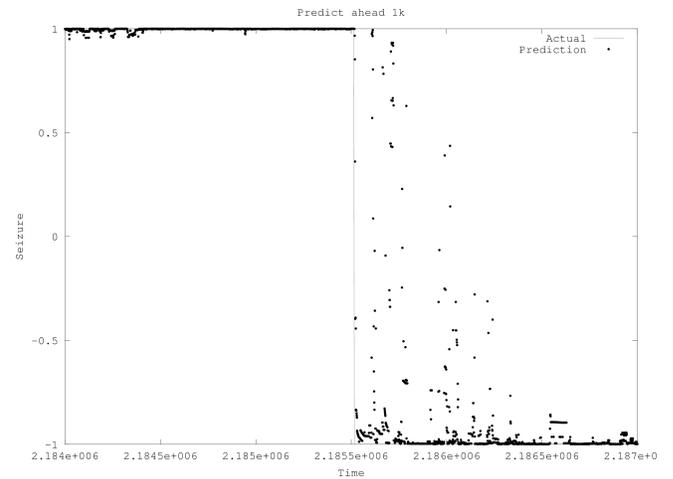

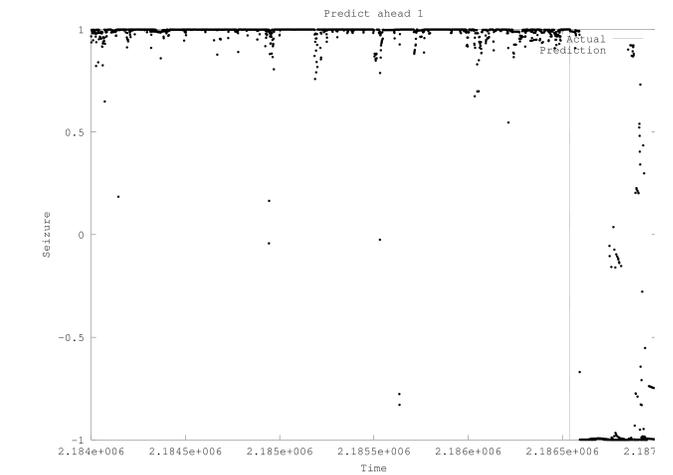

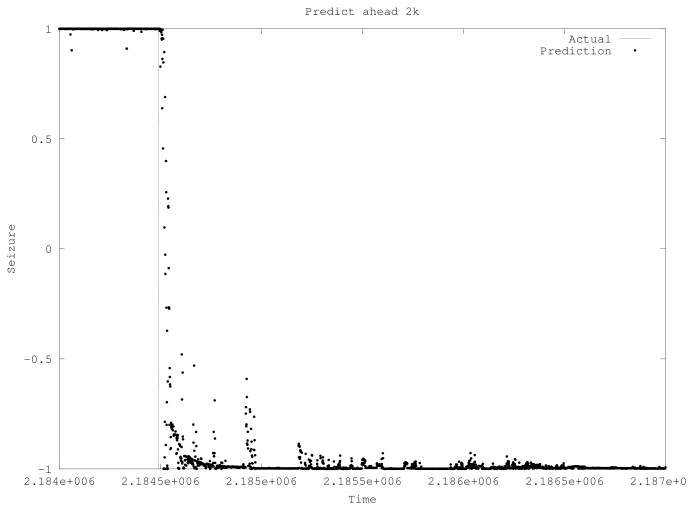

Figure 21: 2048 instants ahead prediction of seizure subsiding

## VI. CONCLUSION

We presented the time-series prediction problem and critically discussed several aspects. Some of the aspects pertain to applications with stringent requirements such as Epileptic Seizure onset prediction. We discussed suitability of Recurrent Neural Networks for time series prediction. We discuss an architecture for Recurrent Neural Networks that grows along three dimensions – number of units in layers, number of hidden layers and extent in number of instants of back-propagation. We highlighted the similarity between RNN with greater than 1 instant back-propagation extent and higher order Markov processes. We prove analytically the fact that additional hidden layers improve the approximation error rate. Application of theory of back-propagation gradient descent leads to a training method called Gradient Descent Back-propagation Through Time-and-Space. We derive a Dynamic Programming (DP) procedure employing matrix operations for the training. DP and use of matrix operations makes the procedure efficient particularly when using data-parallel libraries and on data-parallel or scalar architectures. The simulations studies present the geometry of the parameter space and verify using synthetic data that the network learns the temporal structure in the data so that parameters converge while model output displays same dynamic behavior as the system. Simulation further showed the method to attain near-perfect Average Detection Ratio on real epileptic seizure data.

We have several related ongoing work that are enhancements to reduce cost of software and Integrated Circuit implementations while preserving the very good performance on the discussed as well as other applications. The enhancements fit number of hidden layers, number of units in hidden layers and extent of back-propagation to problem and consider local minima avoidance in gradient descent training. Applications include Natural Language Processing, Automatic Speech Recognition and music analysis.

## APPENDIX

Output of visible node $p$

$$y_{v_p} = f\left(\sum_q w_{v_p u_q} \tilde{y}_{u_q}\right) \quad (33)$$

where $\tilde{y}_{u_0}$ is set to 1, $\tilde{y}_{u_q}$ is the $(q-1)$th input and $w_{v_p u_q}$ is the weight from input $q$ to the visible node $p$. Hence

$$\frac{\partial y_{v_p}}{\partial y_{u_q}} = f'(a)\Big|_{a=a_{v_p}} w_{v_p u_q} \quad (34)$$

and

$$\frac{\partial y_{v_p}}{\partial w_{v_p u_q}} = f'(a)\Big|_{a=a_{v_p}} y_{u_q}. \quad (35)$$

Alternatively, above can be written in matrix notation as

$$\mathbf{y}_v = f(\mathbf{W}_{vu}\tilde{\mathbf{y}}_u), \quad (36)$$

$$\frac{\partial \mathbf{y}_v}{\partial \mathbf{y}_u} = f'(\mathbf{a})\Big|_{\mathbf{a}=\mathbf{a}_v} \mathbf{W}_{vu} \text{ and} \quad (37)$$

$$\frac{\partial \mathbf{y}_v}{\partial \mathbf{W}_{vu}} = f'(\mathbf{a})\Big|_{\mathbf{a}=\mathbf{a}_v}. \quad (38)$$

Now $\mathbf{y}_u$ is the vector of inputs to the network, $\tilde{\mathbf{y}}_u$ is the vector $\mathbf{y}_u$ with a 1 as the additional and the first element (corresponding to the bias input), $\mathbf{W}_{vu}$ is the matrix of weights to visible layer units from the inputs, $f'(\mathbf{a})\Big|_{\mathbf{a}=\mathbf{a}_v}$ is the (diagonal) matrix of derivatives of vector $f(\mathbf{a})$ wrt vector $\mathbf{a}$ evaluated at $\mathbf{a}=\mathbf{a}_v$ and $\partial \mathbf{y}_v / \partial \mathbf{y}_u$ is the matrix of partial derivatives of outputs of visible layer units wrt inputs. $\partial \mathbf{y}_v / \partial \mathbf{W}_{vu}$, derivative of vector $\mathbf{y}_v$ wrt matrix $\mathbf{W}_{vu}$ is also a matrix (and not a 3-dimensional array) as derivative of the output of a unit in a layer wrt weights to other units in the same layer are 0.

Note that representing the set of derivatives as matrices is more than just a matter of compactness. Representing as matrices eases implementation on scalar processors using mathematical library functions by reducing the overheads associated with library function calls and on vector processors such as GPUs enables gains from the data parallelism. However in the rest of the following derivation, for ease of understanding, we continue to use scalar notation and employ sum over elements of vector as necessary.

Output of hidden layer $N_L$ node $j$

$$y_{h_{N_L,j}} = f\left(\sum_{p=1}^{N_H} w_{h_{N_L,j} h_{N_L,p}^-} y_{h_{N_L,p}^-} + \sum_{p=1}^{N_V} w_{h_{N_L,j} v_p} y_{v_p}\right) \quad (39)$$

where $y_{h_{N_L,j}}$ is the output of hidden layer $N_L$ node $j$, $w_{h_{N_L,j} h_{N_L,p}^{(-1)}}$ the weight from hidden layer $N_L$ node $p$ at the previous instant to hidden layer $N_L$ node $j$ at present instant, $y_{h_{N_L,p}^{(-1)}}$ is the output of the hidden layer $N_L$ node $p$ at the previous instant, $w_{h_{N_L,j} v_p}$ the weight from the hidden layer $N_L$ node $j$ to visible node $p$ and $y_{v_p}$ the output of visible node $p$.

Hence

$$\frac{\partial y_{h_{N_L,j}}}{\partial y_{h_{\bar{N}_L,p}}} = f'(a)\Big|_{a=a_{h_{N_L,j}}} w_{h_{N_L,j} h_{\bar{N}_L,p}}, \quad (40)$$

$$\frac{\partial y_{h_{N_L,j}}}{\partial w_{h_{N_L,j} h_{\bar{N}_L,p}}} = f'(a)\Big|_{a=a_{h_{N_L,p}}} y_{h_{\bar{N}_L,p}} \text{ and} \quad (41)$$

$$\frac{\partial y_{h_{N_L,j}}}{\partial w_{h_{N_L,j} v_p}} = f'(a)\Big|_{a=a_{h_{N_L,j}}} y_{v_p}. \quad (42)$$

From (33) and (39),

$$y_{h_{N_L,j}} = f\left(\sum_{p=1}^{N_H} w_{h_{N_L,j} h_{N_L,p}^{(-1)}} y_{h_{N_L,p}^{(-1)}} + \sum_{p=1}^{N_V} w_{h_{N_L,j} v_p} f\left(\sum_{q=1}^{N_I} w_{v_p u_q} y_{u_q}\right)\right) \quad (43)$$

Hence

$$\frac{\partial y_{h_{N_L,j}}}{\partial w_{v_p u_q}} = f'(a)\Big|_{a=a_{h_{N_L,j}}} w_{h_{N_L,j} v_p} f'(a)\Big|_{a=a_{v_p}} y_{u_q}$$

$$= f'(a)\Big|_{a=a_{h_{N_L,j}}} w_{h_{N_L,j} v_p} \frac{\partial y_{v_p}}{\partial w_{v_p u_q}}$$

$$\frac{\partial y_{h_{N_L,j}}}{\partial y_{u_q}} = f'(a)\Big|_{a=a_{h_{N_L,j}}} \sum_{p=1}^{N_V} w_{h_{N_L,j} v_p} f'(a)\Big|_{a=a_{h_{N_L,p}}} w_{v_p u_q}$$

$$= f'(a)\Big|_{a=a_{h_{N_L,j}}} \sum_{p=1}^{N_V} w_{h_{N_L,j} v_p} \frac{\partial y_{v_p}}{\partial y_{u_q}} \quad (44)$$

Output of hidden layer $N_L - 1$,

$$y_{h_{N_L-1,j}} = f\left(\sum_{p=1}^{N_H} w_{h_{N_L-1,j} h_{\bar{N}_L-1,p}} y_{h_{\bar{N}_L-1,p}} + \chi\right) \quad (45)$$

where

$$\chi = \sum_{p=1}^{N_H} w_{h_{N_L-1,j} h_{N_L,p}} f(\xi)$$

$$\xi = \sum_{q=1}^{N_H} w_{h_{N_L,p} h_{\bar{N}_L,q}} y_{h_{\bar{N}_L,q}} + \sum_{q=1}^{N_V} w_{h_{N_L,p} v_q} f\left(\sum_{r=1}^{N_I} w_{v_q u_r} y_{u_r}\right) \quad (46)$$

Hence

$$\frac{\partial y_{h_{N_L-1,j}}}{\partial w_{v_p u_r}}$$

$$= f'(a)\Big|_{a=a_{h_{N_L-1,j}}} \cdot \sum_{p=1}^{N_H} w_{h_{N_L-1,j} h_{N_L,p}} f'(a)\Big|_{a=a_{h_{N_L,p}}} w_{h_{N_L,p} v_q} \frac{\partial y_{v_q}}{\partial w_{v_p u_r}} \quad (47)$$

$$\frac{\partial y_{h_{N_L-1,j}}}{\partial w_{v_p u_r}} = f'(a)\Big|_{a=a_{h_{N_L-1,j}}} \sum_{p=1}^{N_H} w_{h_{N_L-1,j} h_{N_L,p}} \frac{\partial y_{h_{N_L,p}}}{\partial w_{v_p u_r}} \quad (48)$$

(23) and (26) can in fact be generalized so we may compute the derivative of a unit output or a weight wrt to a unit output (say $y_{x_1}$) at an arbitrary distance away in terms of the derivative of the same weight or output wrt to (an) output(s) $y_{x_2}$ distance 1 away, the weight connecting $y_{x_2}$ to $y_{x_1}$ and derivative of the original unit output or weight wrt $y_{x_2}$. For example in a network that has $N_L$ hidden layers, we are back propagating $T$ instants, the derivative of final output $y_{z_i}$ wrt the weight associated with input $q$ at instant $-T$ to input layer unit $p$ is given by

$$\frac{\partial y_{z_i}}{\partial w_{v_p^{(-T)} u_q^{(-T)}}} = f'(a)\Big|_{a=a_{z_i}} \sum_{p=1}^{N_H} w_{z_i h_{N_L,p}} \frac{\partial y_{h_{N_L,p}}}{\partial w_{v_p^{(-T)} u_q^{(-T)}}} \quad (49)$$

Above suggests a dynamic programming method for computing the derivatives comprising of traversing the graph from inputs to outputs, computing derivative of unit outputs wrt distance 1 away unit outputs, distance 1 away weights and distance 2 away weights. Derivatives wrt unit outputs at distance greater than 1 and weights at distance greater than 2 are computed by taking weighted sums over corresponding derivatives of distance 1 away unit outputs. The distances here are partially over space and partially over time.

In the joint-state-and-parameter-vector $\mathbf{x}_k$, the individual layer unit outputs and unit input weights are assigned indices so as to facilitate computation of the derivatives. $1 \leq i \leq N_O$ are the current final outputs, $N_O + 1 \leq i \leq N_O + N_L N_H$ are the current hidden layer unit outputs, $N_O + N_L N_H + 1 \leq i \leq N_O + N_L N_H + N_V = N_U$ are the current visible layer unit outputs. Current outputs are followed by instant $-1$, instant $-2$, …, instant $-T$ outputs where $T$ is the extent of back-propagation. Weights follow the unit outputs. Current weights associated with output layer units occupy indices

$$(T+1)N_U + 1 \leq i \leq (T+1)N_U + N_O(1 + N_H), \quad (50)$$

those with hidden layers

$$(T+1)N_U + N_O(1 + N_H) + 1 \leq i$$
$$\leq (T+1)N_U + N_O(1 + N_H) + N_H(1 + N_V + N_H)$$
$$+ (N_L - 1)N_H(1 + 2N_H) \quad (51)$$

and finally those with the input layer

$$(T+1)N_U + N_O(1 + N_H) + N_H(1 + N_V + N_H)$$
$$+ (N_L - 1)N_H(1 + 2N_H) + 1 \leq i$$
$$\leq (T+1)N_U + N_O(1 + N_I) + N_H(1 + N_V + N_H)$$
$$+ (N_L - 1)N_H(1 + 2N_H) + N_V(1 + N_I)$$
$$= (T+1)N_U + N_W. \quad (52)$$

As with outputs, current weights are followed by instant $-1$ through instant $-T$ weights.

Before computing new current outputs, the current through instant $-T+1$ outputs are copied to instant $-1$ through instant $-T$ outputs. Similarly before updating, the current through instant $-T+1$ weights are copied to instant $-1$ through instant $-T$ weights.

However significant time-efficiency is achieved by noting that no actual copying needs to be performed. The unit outputs and weights sections are treated as $T+1$ (instants) long circular buffers and the current instant pointer is simply advanced one instant!

The row and column indices of matrices of derivatives of unit outputs and weights wrt upstream unit outputs and weights are also similarly incremented circularly when advancing to next instant. This makes it unnecessary to physically copy weights. As operations of matrix-vector product and vector-vector inner product involves summing over products of pairs of elements and the sum operation is order-independent, the matrix operations are not affected. This makes the time-complexity when using back-propagation extent greater than 1 same as the time complexity when using back-propagation extent equal to 1.

With $N_O$ long network outputs vector $\mathbf{y}_z$ and the derivatives of weights wrt network outputs arranged in a $N_O \times N_W$ matrix $\mathbf{J}$, the updated weights are given by $\mathbf{w} \leftarrow \mathbf{w} - \eta \mathbf{J}(\hat{\mathbf{y}}_z - \mathbf{y}_z)$.

Thus, the space complexity is $O((T+1)(N_U + N_W))$. With $T$ a constant independent of network dimensions, this is linear in network dimension. With access to previous outputs and weights requiring $O(1)$ time, the time-complexity of computations at each instant is $O(2N_H N_W)$ under normally valid assumptions of $N_H > N_I$ and $N_H > N_O$.

ACKNOWLEDGMENT

Piyush was a summer intern at the Nano-Electronics Research Laboratory in the Electrical Engineering Dept of Purdue University for a part of this work and is grateful for support and encouragement received from Prof K. Roy and Dr S. Raghunathan.

**Sharat C Prasad** (M'87) received the B.Tech. degree in Electronics and Communication Engineering from I.I.T. Kaharagpur, India in 1983, M.E. in Computer Science and Automation from I.I.Sc. Bangalore India in 1987 and Ph.D. in Electrical Engineering from University of Texas at Dallas in 2005.

He was an Intern at the Tata Institute of Fundamental Research during the summer of 1982, worked at the Indian Subsidiary of International Computers Limited, U.K. from 1983 until 1985, at Texas Instruments from 1987 until 1997, at Cisco Systems from 1998 until 2006, at Google Inc from 2007 until 2012 and is with Best3 Corp since 2012.

Dr Prasad is a member of IEEE and ACM. He has authored or co-authored one book, over 5 journal, over 10 conference and over 30 patents.

**Piyush Prasad** was born in Plano, Texas, in 1995. He is expecting to receive his B.S. degree in Biomedical Engineering from Washington University, St Louis, MO in 2017.

He was a summer intern at the Nano-electronics laboratory of Dept of Electrical Engineering at Purdue University.

Mr Prasad was a recipient of several awards at a national Robotics league and is the co-author of a patent.